%% file: main.tex
\pdfoutput=1

\documentclass[11pt]{article}

\usepackage[final]{acl2025}

\usepackage{times}
\usepackage{latexsym}

\usepackage[T1]{fontenc}

\usepackage[utf8]{inputenc}

\usepackage{microtype}

\usepackage{inconsolata}

\usepackage{graphicx}

\usepackage{color}
\usepackage{booktabs}
\usepackage{multirow}
\usepackage{colortbl}
\usepackage{caption}
\usepackage{subfigure}
\usepackage{graphicx}
\usepackage{hyperref}
\usepackage{amsfonts,amssymb}
\usepackage{longtable}
\usepackage[most]{tcolorbox}

\title{M\textsc{otive}B\textsc{ench}: How Far Are We From Human-Like Motivational Reasoning in Large Language Models?}

\author{Xixian Yong\textsuperscript{\rm 1} \quad Jianxun Lian\textsuperscript{\rm 4} \quad Xiaoyuan Yi\textsuperscript{\rm 4} \quad Xiao Zhou\textsuperscript{\rm 1,2,3}\thanks{Corresponding author.} \quad Xing Xie\textsuperscript{\rm 4} \\
\textsuperscript{\rm 1}Gaoling School of Artificial Intelligence, Renmin University of China \\
\textsuperscript{\rm 2}Beijing Key Laboratory of Research on Large Models and Intelligent Governance \\
\textsuperscript{\rm 3}Engineering Research Center of Next-Generation Intelligent Search and Recommendation, MOE \\
\textsuperscript{\rm 4}Microsoft Research Aisa \\
\texttt{\{xixianyong,xiaozhou\}@ruc.edu.cn}}

\begin{document}
\maketitle
\begin{abstract}
Large language models (LLMs) have been widely adopted as the core of agent frameworks in various scenarios, such as social simulations and AI companions. However, the extent to which they can replicate human-like motivations remains an underexplored question. Existing benchmarks are constrained by simplistic scenarios and the absence of character identities, resulting in an information asymmetry with real-world situations. To address this gap, we propose M\textsc{otive}B\textsc{ench}, which consists of 200 rich contextual scenarios and 600 reasoning tasks covering multiple levels of motivation. Using M\textsc{otive}B\textsc{ench}, we conduct extensive experiments on seven popular model families, comparing different scales and versions within each family. The results show that even the most advanced LLMs still fall short in achieving human-like motivational reasoning. Our analysis reveals key findings, including the difficulty LLMs face in reasoning about \textit{"love \& belonging"} motivations and their tendency toward excessive rationality and idealism. These insights highlight a promising direction for future research on the humanization of LLMs. The dataset, benchmark, and code are available at \url{https://aka.ms/motivebench}.
\end{abstract}

\section{Introduction}

\begin{figure}[t]
\centering
\includegraphics[width=1\linewidth]{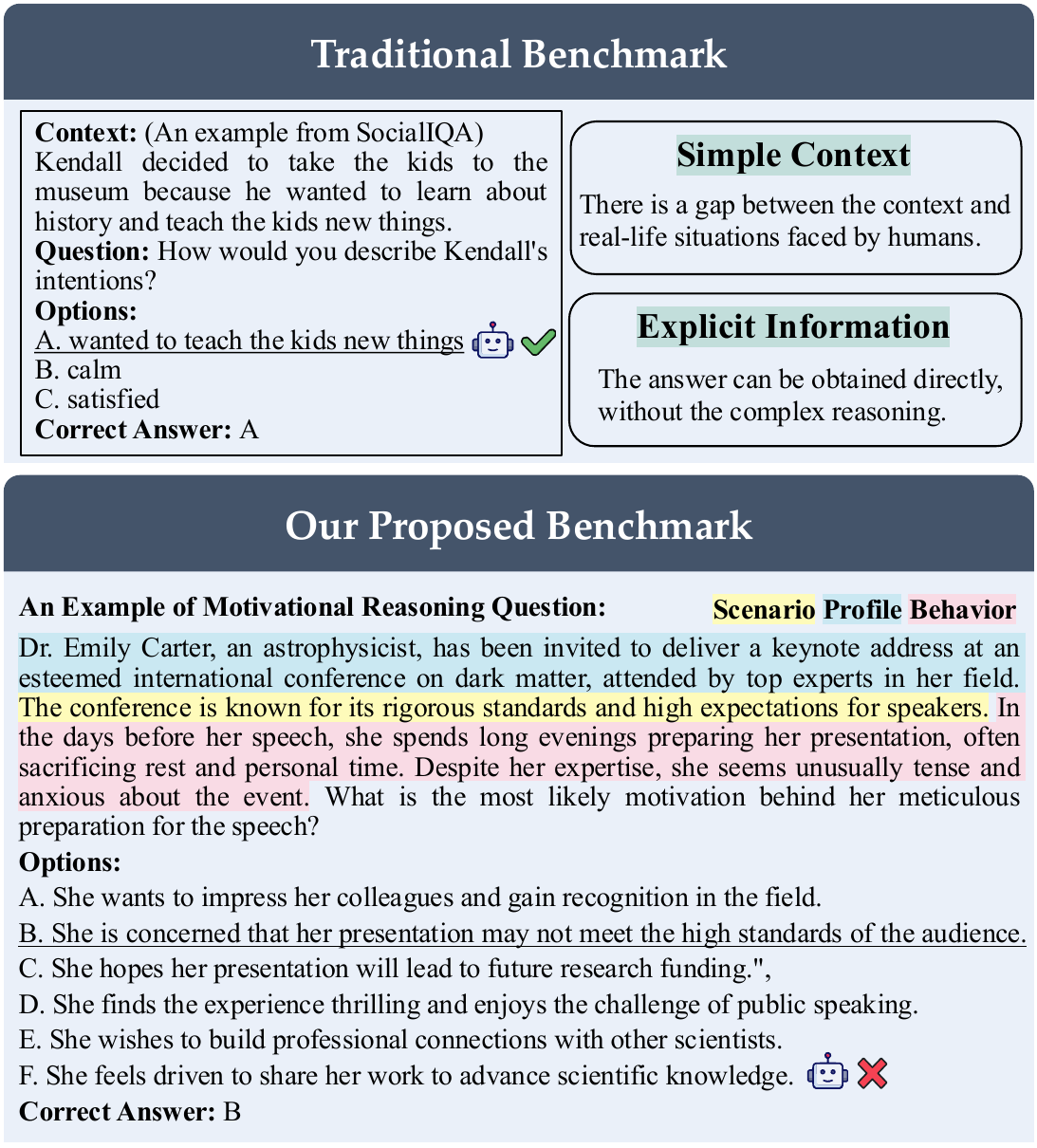}
\caption{The difference between the existing traditional benchmark, such as S\textsc{ocial}IQ\textsc{a} from~\citet{sap2019socialiqa} and our proposed M\textsc{otive}B\textsc{ench}.}
\label{compare}
\vspace{-1.5em}
\end{figure}

\begin{figure*}[t]
\centering
\includegraphics[width=1\textwidth]{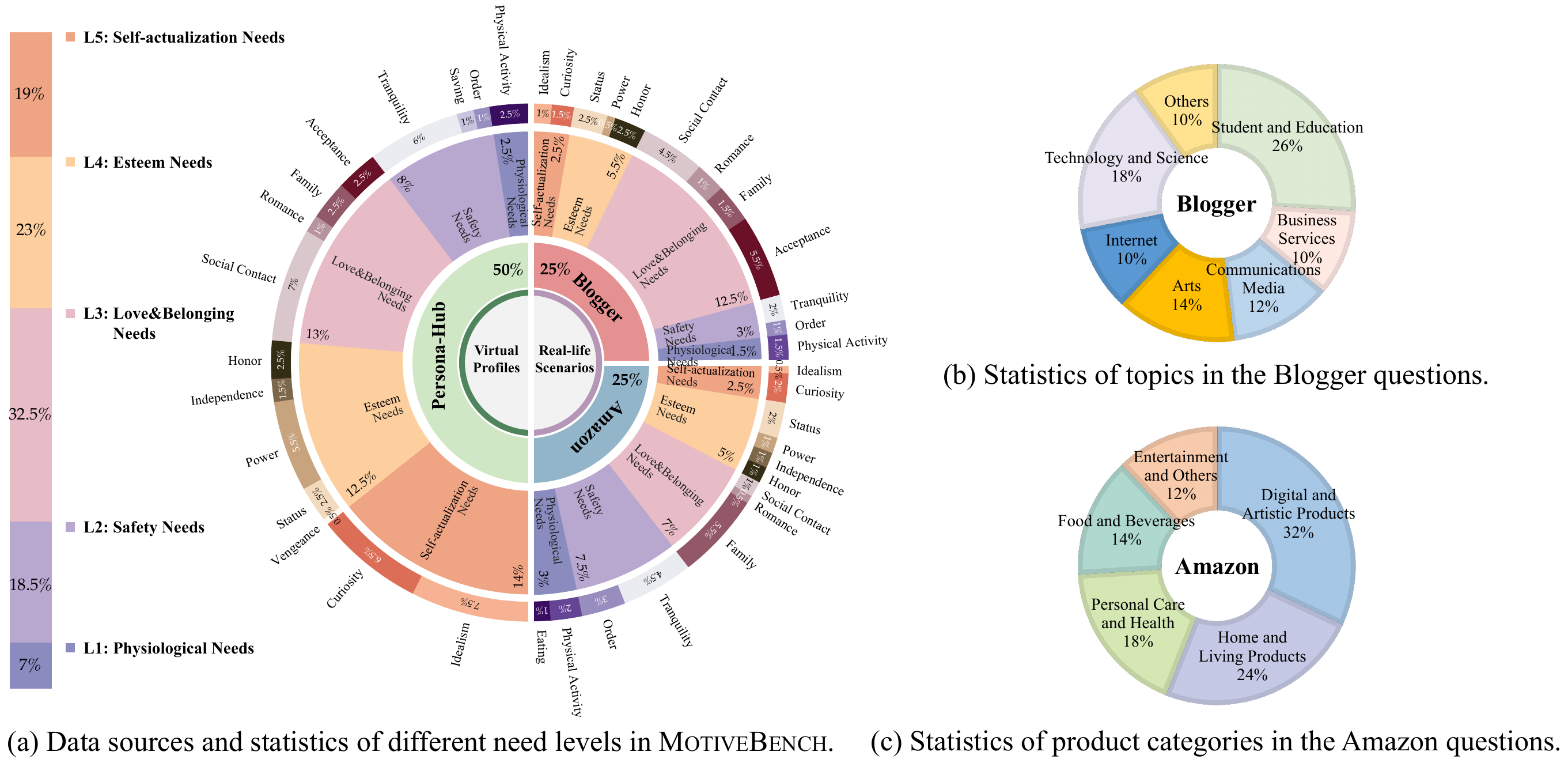}
\caption{The statistical overview of M\textsc{otive}B\textsc{ench}. It contains 200 diverse profiles and real-world scenarios, along with 600 motivational and behavioral reasoning questions, covering multiple finely-grained levels of needs.}
\label{data_statistics}
\vspace{-0.8em}
\end{figure*}

Motivation is commonly conceptualized as an internal drive or psychological force that influences individuals to initiate and sustain goal-oriented activities~\cite{hagger2005social,brehm2014psychology}. It serves as a key explanatory factor for understanding why people initiate, continue, or terminate specific behaviors at any given scenarios~\cite{kazdin2000encyclopedia}. Motivation can be intrinsic, driven by internal values or preferences, or extrinsic, shaped by external rewards or punishments~\cite{ryan2000self,radel2016does}.

Mimicking human behavior in specific scenarios has been a crucial task for autonomous agents, forming the foundation for various applications such as problem-solving, testing, and simulation~\cite{schatzmann2007agenda}. Previous studies employ either rule-based~\cite{keizer2010parameter,wilkins2014practical} or machine learning approaches~\cite{asri2016sequence,kreyssig2018neural} to replicate human interactions in isolated and controlled environments. 

With the advent of large language models (LLMs) like GPT-4~\cite{achiam2023gpt} and LLaMA~\cite{dubey2024llama}, an increasing number of studies have been adopting LLM-based agents~\cite{aher2023using,argyle2023out,boiko2023emergent,kang2023chatmof,mehta2023improving,hong2023metagpt,wang2024survey} due to the remarkable capabilities of LLMs in general problem-solving, reasoning, and autonomous action-taking. However, a critical question remains underexplored: \textbf{Can current LLMs truly understand and exhibit human-like motivations and behaviors?} The complexity of human behavior dynamics poses new challenges for LLMs, which are distinct from the challenges in understanding and manipulating the physical world. Delving deeply into this research question can lay a reliable foundation for LLM-based social simulations.

Recent literature has attempted to explore the gap between LLMs and real humans from a narrow behavioral perspective. \citet{xie2024can} demonstrated the feasibility of simulating human trust behavior with LLM agents in the Trust Game~\cite{berg1995trust}, while \citet{zhou2023sotopia} analyzed differences between LLMs and humans in social interactions~\cite{lee2024towards}. However, these studies focus on specific subdomains and fail to provide broader insights into human-like behavior. In the study of motivations and behaviors in LLMs, \citet{sap2019socialiqa} introduced S\textsc{ocial}IQ\textsc{a}, a benchmark for commonsense reasoning in social contexts. While it includes some basic reasoning tasks on intentions behind behaviors, it lacks detailed, comprehensive, and challenging assessments, as shown in Figure~\ref{compare}. Existing similar benchmarks~\cite{rashkin2018modeling,rashkin2018event2mind,talmor2018commonsenseqa} also exhibit several clear limitations: \textbf{(1) Simplistic contexts}, lacking detailed scenarios and character profiles, leading to information asymmetry compared to real-world situations; \textbf{(2) Overly explicit information}, with tasks solvable through basic pattern matching without requiring human-like reasoning; and \textbf{(3) Limited theoretical grounding}, failing to systematically capture the multi-level nature of human motivation.

To address the above challenges, we propose M\textsc{otive}B\textsc{ench}, a comprehensive evaluation benchmark, consisting of 200 diverse profiles and scenarios, along with 600 motivational and behavioral reasoning questions. Figure~\ref{data_statistics} shows an overview of M\textsc{otive}B\textsc{ench}. We strive to cover and balance the proportions of different levels of needs in the benchmark, ultimately formulating the questions. Specifically, it has the following advantages: \textbf{(1) Diverse scenarios, profiles, motivations, and behaviors.} We utilize diverse profiles from the Persona-Hub~\cite{ge2024scaling} dataset, along with real-world motivation and behavior data from platforms like Amazon and Blogger, as the basis for question generation.  \textbf{(2) Human-in-the-loop multi-agent framework to enhance efficiency and quality.} We propose a multi-agent collaboration framework that efficiently generates high-quality questions across a range of difficulties, requiring minimal human effort to ensure validity. \textbf{(3) Grounded in authoritative psychological theories to ensure comprehensive evaluation.} Our test questions cover the five levels of Maslow's Hierarchy of Needs~\cite{maslow1943theory}, as well as the 16 basic desires of human nature from the Reiss Motivation Profile~\cite{reiss2004multifaceted}.

To the best of our knowledge, M\textsc{otive}B\textsc{ench} is the first benchmark for evaluating LLMs' human-like motivation-behavior reasoning. \textbf{Our results show that even the most advanced LLMs fall short in achieving human-like motivation and behavior reasoning.} Beyond the quantified results, comprehensive experiments reveal notable insights, such as significant differences in motivational reasoning between LLMs and humans, and their limitations in data annotation for human social behaviors. We hope our research provides practical guidelines for applying LLMs in social simulations and contributes to their future humanization.

\begin{figure*}[t]
\centering
\includegraphics[width=1\textwidth]{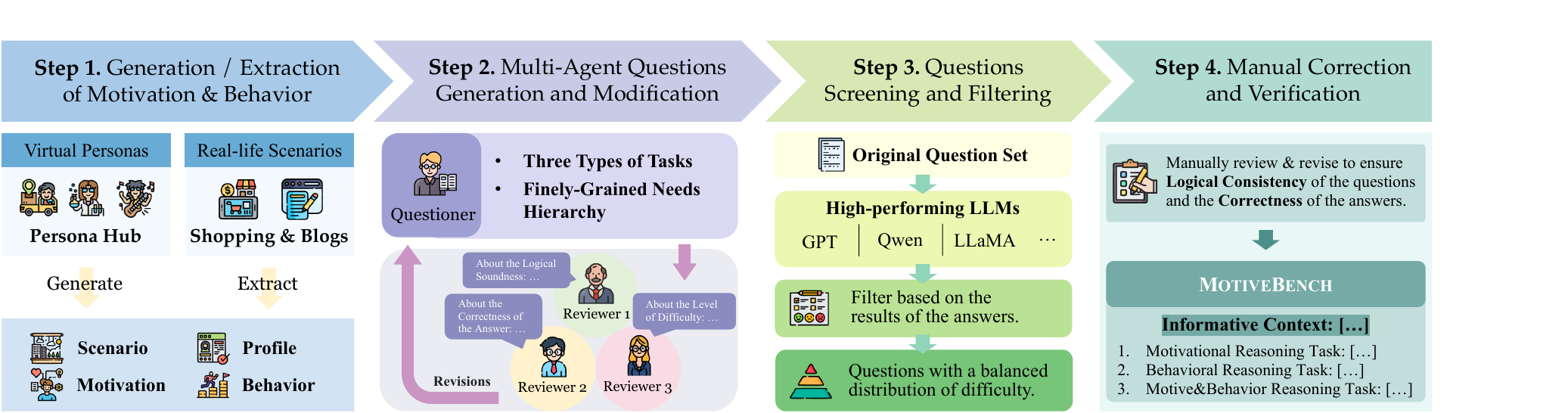}
\caption{A step-by-step questions generation and correction pipeline using AI-Human collaboration framework.}
\label{pipeline}
\end{figure*}

\section{M\textsc{otive}B\textsc{ench} Preliminaries}
\subsection{Three Types of Reasoning Tasks}
Generally, in a specific scenario, an individual with a certain profile will perform a behavior based on a particular motivation. We define this as a complete behavioral quadruple: \textbf{Scenario}, \textbf{Profile}, \textbf{Motivation}, and \textbf{Behavior}. The scenario provides the context and external triggers~\cite{yang2009concept}. Profile shapes an individual's understanding of the behavior and the way they act~\cite{bandura2001social,eagly2012social}. Motivation is the internal driving force behind an individual's actions, based on their needs, goals, or emotional state~\cite{deci2012self,shayganfar2016overview}. Behavior is the result of the interaction between scenario, profile, and motivation~\cite{graham1991review}. Therefore, we define three types of reasoning tasks:

1) \textbf{Motivational Reasoning Question.} Given a specific scenario, profile, and detailed behavior information, the task is to infer the motivation behind the individual's behavior.

2) \textbf{Behavioral Reasoning Question.} Given a specific scenario, profile, and detailed motivation information, the task is to infer the most likely behavior the individual would perform.

3) \textbf{Motive\&Behavior  Reasoning Question.} This more challenging task closely aligns with the ultimate goal of autonomous agents, which is to infer the most reasonable motivation and behavior when only the scenario and profile are provided.

\subsection{Fine-Grained Needs Hierarchy}
\label{sec.needs}
Maslow's hierarchy of needs theory~\cite{maslow1943theory} suggests that human actions are driven by various needs, which are divided into five levels: Physiological Needs, Safety Needs, Love and Belonging Needs, Esteem Needs, and Self-actualization Needs. When lower-level needs are met, individuals will seek to fulfill higher-level needs~\cite{jr1991becoming}.

Furthermore, \citet{reiss2004multifaceted} proposes 16 more granular categories to provide a broader and more informative range of motivations. These include Curiosity, Idealism, Honor, Independence, Power, Status, Vengeance, Acceptance, Family, Romance, Social Contact, Order, Saving, Tranquility, Eating, and Physical Activity. Although the Reiss theory offers more detailed insights into motivation, the broader range of abstract concepts can be difficult  to manage. Inspired by \citet{rashkin2018modeling}, we adopt a hybrid method in which the Reiss Motive Profile labels are categorized as sub-categories within Maslow's framework.

\section{M\textsc{otive}B\textsc{ench} Construction}
We construct M\textsc{otive}B\textsc{ench} from scratch to avoid relying on existing scales or test items from psychology or sociology, thereby mitigating potential data leakage or contamination issues. Figure~\ref{pipeline} illustrates the AI-Human collaboration framework.

\subsection{Data Collection and Pre-processing}
To obtain diverse scenarios and profiles, we collect and pre-process data from Persona-Hub\footnote{https://github.com/tencent-ailab/persona-hub} proposed by~\citet{ge2024scaling}, as well as real-life platforms like Amazon\footnote{https://amazon-reviews-2023.github.io/}~\cite{hou2024bridging} and Blogger\footnote{https://www.kaggle.com/datasets/rtatman/blog-authorship-corpus }~\cite{schler2006effects}.

Persona-Hub contains diverse profiles, such as \textit{"A fellow astrophysicist who specializes in the study of dark matter and provides valuable insights and critiques to the author's research."} Based on them, we can synthesize a diverse range of scenarios, motivations, and behaviors using the fine-grained hierarchy of needs outlined in Section~\ref{sec.needs}.

\begin{figure}[t]
\centering
\includegraphics[width=1\linewidth]{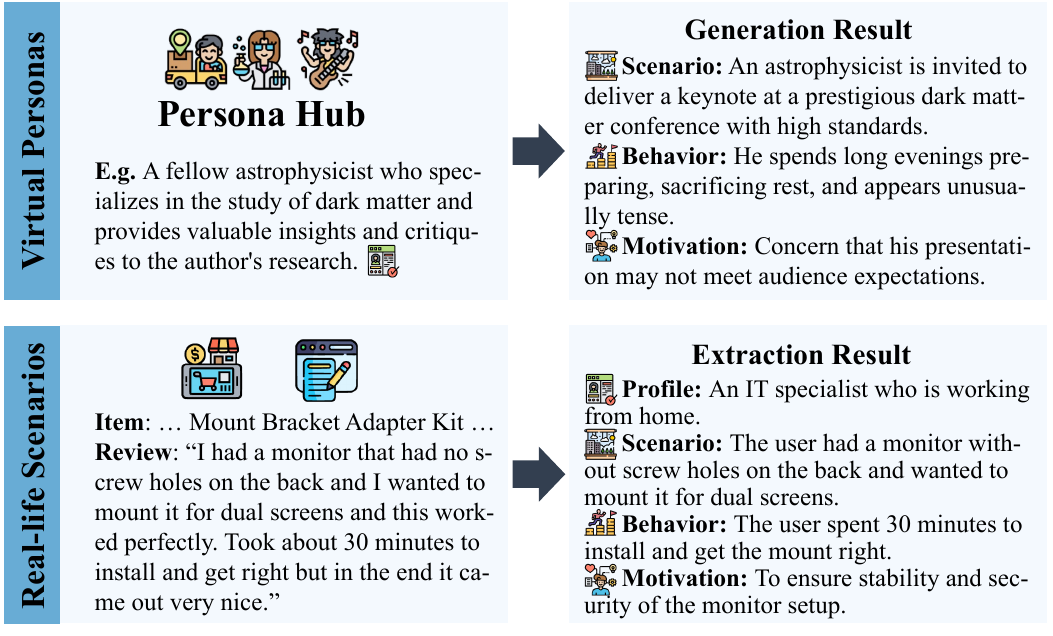}
\caption{Generation and extraction of motivations and behaviors from virtual personas and real-life scenarios.}
\label{data_collection}
\vspace{-1em}
\end{figure}

In addition, review texts on platforms like Amazon offer rich data reflecting genuine user intent, such as specific reasons for purchasing a product. Blog posts from Blogger, composed of discrete and often informal diary-style entries, also serve as valuable sources. These texts provide grounded, real-world scenarios that are useful for data collection and analysis. Therefore, we first collect review data from 33 Amazon product domains to ensure diversity. For blog posts, we use the dataset proposed by~\citet{schler2006effects}. We then employ LLMs, such as LLaMA3.1-70B and Qwen2.5-72B, to filter high-quality reviews and posts that align with Maslow's hierarchy of needs, focusing on motivations rooted in real-life contexts. For example, in reviews of smartphones, we prioritize motivations such as purchasing for better communication with family, reflecting "social needs", or selecting a phone with a long battery life for convenience during travel, reflecting "safety needs", rather than only focusing on product attributes (such as performance or appearance) as the motivation for purchase. An example is shown in Figure~\ref{data_collection}.

\subsection{Questions Generation and Modification}
Existing benchmarks mostly rely on manual construction, which requires significant effort~\cite{sabour2024emobench,chen2024tombench}. To reduce financial and labor costs, we employ a multi-agent framework to formulate questions and generate options.

Specifically, we begin by using an LLM-based questioner to formulate complete question content based on the quadruples obtained in the previous stage. Next, three LLM-based reviewers provide feedback from: 1) the logical soundness of the question, 2) the correctness of the answer, and 3) the level of difficulty. The feedback is then compiled and sent back to the questioner for revisions. This process is iterated until no further suggestions are made, or the iteration threshold is reached. To minimize the hidden biases introduced by a single LLM, we use a diverse set of models, such as LLaMA3.1-70B and Qwen2.5-72B, for question creation and modification. The detailed prompts of our agents can be found in the Appendix~\ref{sec:prompts}.

\subsection{Questions Screening and Filtering}
A well-rounded benchmark should include questions of varying difficulty to evaluate LLMs. Since controlling difficulty during question generation is challenging, we use high-performing LLMs to answer and label each question. Based on this, we categorize the questions into difficult, medium, and easy levels, ensuring a balanced distribution of all three difficulty levels in the final benchmark. Ultimately, We curate a subset of high-quality, diverse scenarios, retaining 100 scenarios for Person-Hub, 50 for Amazon, and 50 for Blogger. These scenarios span different difficulty levels and provide balanced coverage. Notably, this scale of context-rich evaluation proves sufficient for assessing LLMs' performance, yielding statistically robust and methodologically sound conclusions~\cite{sabour2024emobench}.

\subsection{Manual Correction and Verification}
After filtering the questions, we authors manually review and revise each question to ensure high quality. We carefully refine the logical consistency of questions and the consensus of the correct options, it taks about 6 minutes per question. Due to the hallucination issues~\cite{huang2023survey} in LLMs that often introduce logical or factual errors, this step is essential. Our analysis reveals that some generated questions lack clear human consensus, highlighting LLMs' limitations in capturing human reasoning's complexity and nuance. Notably, we minimize the impact of demographic backgrounds on answers and demonstrate this by recruiting diverse annotators in follow-up experiments. The final statistics of M\textsc{otive}B\textsc{ench} are shown in Table~\ref{data_statistics_2}, where the context length is significantly longer than that of previous benchmarks (e.g., S\textsc{ocial}IQ\textsc{a} in~\citet{sap2019socialiqa}: 20.16 tokens per context, and C\textsc{ommonsense}QA in~\citet{talmor2018commonsenseqa}: 13.41 tokens per context).

\begin{table}[htbp]
  \centering
  \resizebox{\linewidth}{!}{
    \begin{tabular}{lccccc}
    \toprule
          & \multicolumn{1}{c}{\multirow{2}[4]{*}{\textbf{\#S}}} & \multicolumn{1}{c}{\multirow{2}[4]{*}{\textbf{\#Q}}} & \multicolumn{3}{c}{\textbf{Average Token Count}} \\
    \cmidrule{4-6}          
          &       &       & \textbf{Context} & \textbf{Cor.Opt} & \textbf{Fal.Opt} \\
    \midrule
    \textbf{M\textsc{otive}B\textsc{ench}} & \textbf{200}   & \textbf{600}   & \textbf{96.45}  & \textbf{13.83}  & \textbf{13.31}  \\
    \midrule
    Persona & 100   & 300   & 97.13  & 15.53  & 14.60  \\
    Amazon & 50    & 150   & 87.19  & 12.57  & 12.51  \\
    Blogger  & 50    & 150   & 224.68  & 21.32  & 21.41  \\
    \bottomrule
    \end{tabular}}
    \caption{Data statistics of our proposed M\textsc{otive}B\textsc{ench}. (\#S: number of scenarios, \#Q: number of questions, Cor.Opt: correct options, and Fal.Opt: false options.)}
  \label{data_statistics_2}
  \vspace{-1em}
\end{table}%

Finally, we recruit 28 participants and select 15 (diverse backgrounds) to assess MCQ correctness and reasonableness while preventing answer leakage. Each question is independently annotated by five annotators, yielding 98.17\% consistency ($\geq 3$) and 93.00\% agreement with predefined answers. Detailed analyses are in the Appendix~\ref{annatation}. We correct inconsistent or contradictory questions based on annotations, ensuring each final answer aligns accurately with human consensus.

\begin{table*}[t]
  \centering
  \resizebox{\textwidth}{!}{
    \begin{tabular}{l|cccccc|cc}
    \toprule
    \multicolumn{1}{p{10.915em}|}{\textbf{Motive\&Behavior}} & \multicolumn{2}{c}{\textbf{Vitural Profiles}} & \multicolumn{4}{c|}{\textbf{Real-life Scenarios}} & \multicolumn{2}{c}{\multirow{2}[2]{*}{\textbf{Overall}}} \\
    \textbf{Reasoning Ability} & \multicolumn{2}{c}{\textbf{Persona-Hub}} & \multicolumn{2}{c}{\textbf{Amazon}} & \multicolumn{2}{c|}{\textbf{Blogger}} & \multicolumn{2}{c}{} \\
    \midrule
    \makebox[0.28\textwidth][l]{\textbf{LLMs}} & \makebox[0.095\textwidth][c]{\textbf{Base}} & \makebox[0.095\textwidth][c]{\textbf{CoT}} & \makebox[0.095\textwidth][c]{\textbf{Base}} & \makebox[0.095\textwidth][c]{\textbf{CoT}} & \makebox[0.095\textwidth][c]{\textbf{Base}} & \makebox[0.095\textwidth][c]{\textbf{CoT}} & \makebox[0.095\textwidth][c]{\textbf{Base}} & \makebox[0.095\textwidth][c]{\textbf{CoT}} \\
    \midrule
    \href{https://huggingface.co/baichuan-inc/Baichuan2-7B-Chat}{Baichuan2-7B-Chat} & \cellcolor[rgb]{ .867,  .922,  .969}\textbf{33.61 } & 27.17  & \cellcolor[rgb]{ .867,  .922,  .969}\textbf{40.22 } & 29.78  & \cellcolor[rgb]{ .867,  .922,  .969}\textbf{42.44 } & 35.67  & \cellcolor[rgb]{ .867,  .922,  .969}\textbf{37.47 } & 29.94  \\
    \href{https://huggingface.co/baichuan-inc/Baichuan2-13B-Chat}{Baichuan2-13B-Chat} & 21.78  & \cellcolor[rgb]{ .886,  .937,  .855}\textbf{36.22 } & 36.56  & \cellcolor[rgb]{ .886,  .937,  .855}\textbf{49.11 } & 18.22  & \cellcolor[rgb]{ .886,  .937,  .855}\textbf{37.78 } & 24.59  & \cellcolor[rgb]{ .886,  .937,  .855}\textbf{39.83 } \\
    \midrule
    \href{https://huggingface.co/THUDM/chatglm3-6b}{ChatGLM3-6B} & 29.67  & 26.94  & 29.44  & 39.33  & 52.00  & 37.33  & 35.20  & 32.64  \\
    \href{https://huggingface.co/THUDM/glm-4-9b-chat}{GLM4-9B-Chat} & \cellcolor[rgb]{ .867,  .922,  .969}\textbf{62.72 } & \cellcolor[rgb]{ .886,  .937,  .855}\textbf{53.00 } & \cellcolor[rgb]{ .867,  .922,  .969}\textbf{79.22 } & \cellcolor[rgb]{ .886,  .937,  .855}\textbf{76.22 } & \cellcolor[rgb]{ .867,  .922,  .969}\textbf{66.67 } & \cellcolor[rgb]{ .886,  .937,  .855}\textbf{52.00 } & \cellcolor[rgb]{ .867,  .922,  .969}\textbf{67.83 } & \cellcolor[rgb]{ .886,  .937,  .855}\textbf{58.56 } \\
    \midrule
    \href{https://huggingface.co/01-ai/Yi-1.5-6B-Chat}{Yi1.5-6B-Chat} & 47.89  & 36.67  & 54.33  & 50.44  & 52.22  & 44.89  & 50.58  & 42.17  \\
    \href{https://huggingface.co/01-ai/Yi-1.5-9B-Chat}{Yi1.5-9B-Chat} & 57.78  & 41.89  & 66.56  & 49.78  & 56.89  & 52.67  & 59.75  & 46.56  \\
    \href{https://huggingface.co/01-ai/Yi-1.5-34B-Chat}{Yi1.5-34b-Chat} & \cellcolor[rgb]{ .867,  .922,  .969}\textbf{59.83 } & \cellcolor[rgb]{ .886,  .937,  .855}\textbf{56.72 } & \cellcolor[rgb]{ .867,  .922,  .969}\textbf{78.00 } & \cellcolor[rgb]{ .886,  .937,  .855}\textbf{70.11 } & \cellcolor[rgb]{ .867,  .922,  .969}\textbf{64.56 } & \cellcolor[rgb]{ .886,  .937,  .855}\textbf{56.67 } & \cellcolor[rgb]{ .867,  .922,  .969}\textbf{65.56 } & \cellcolor[rgb]{ .886,  .937,  .855}\textbf{60.06 } \\
    \midrule
    \href{https://huggingface.co/microsoft/Phi-3-mini-4k-instruct}{Phi3-mini-4k-Instruct} & 59.61  & 44.78  & 70.33  & 60.89  & 66.78  & 41.44  & 64.08  & 47.97  \\
    \href{https://huggingface.co/microsoft/Phi-3-small-8k-instruct}{Phi3-small-8k-Instruct} & 64.50  & 55.33  & 80.89  & \cellcolor[rgb]{ .886,  .937,  .855}\textbf{79.00 } & 64.44  & 60.33  & 68.58  & 62.50  \\
    \href{https://huggingface.co/microsoft/Phi-3-medium-4k-instruct}{Phi3-medium-4k-Instruct} & 68.06  & 56.89  & \cellcolor[rgb]{ .867,  .922,  .969}\textbf{84.44 } & 73.56  & \cellcolor[rgb]{ .867,  .922,  .969}\textbf{71.67 } & \cellcolor[rgb]{ .886,  .937,  .855}\textbf{61.22 } & \cellcolor[rgb]{ .867,  .922,  .969}\textbf{73.06 } & 62.14  \\
    \href{https://huggingface.co/microsoft/Phi-3.5-mini-instruct}{Phi3.5-mini-Instruct} & 60.06  & 48.39  & 73.67  & 61.56  & 68.44  & 54.67  & 65.56  & 53.25  \\
    \href{https://huggingface.co/microsoft/Phi-3.5-MoE-instruct}{Phi3.5-MoE-Instruct} & \cellcolor[rgb]{ .867,  .922,  .969}\textbf{70.11 } & \cellcolor[rgb]{ .886,  .937,  .855}\textbf{62.67 } & 76.11  & 76.44  & 69.44  & 59.22  & 71.44  & \cellcolor[rgb]{ .886,  .937,  .855}\textbf{65.25 } \\
    \midrule
    \href{https://huggingface.co/meta-llama/Llama-2-7b-chat-hf}{Llama2-7B-Chat} & 24.78  & 22.67  & 34.11  & 27.33  & 42.00  & 32.44  & 31.42  & 26.28  \\
    \href{https://huggingface.co/meta-llama/Llama-2-13b-chat-hf}{Llama2-13B-Chat} & 45.17  & 36.33  & 53.22  & 42.78  & 58.33  & 44.33  & 50.47  & 39.94  \\
    \href{https://huggingface.co/meta-llama/Llama-2-70b-chat-hf}{Llama2-70B-Chat} & 54.11  & 46.78  & 63.78  & 57.78  & 65.44  & 62.33  & 59.36  & 53.42  \\
    \href{https://huggingface.co/meta-llama/Llama-3.1-8B-Instruct}{Llama3.1-8B-Instruct} & 52.06  & 54.11  & 59.56  & 73.89  & \cellcolor[rgb]{ .867,  .922,  .969}\textbf{69.33 } & 58.00  & 58.25  & 60.03  \\
    \href{https://huggingface.co/meta-llama/Llama-3.1-70B-Instruct}{Llama3.1-70B-Instruct} & \cellcolor[rgb]{ .867,  .922,  .969}\textbf{73.83 } & \cellcolor[rgb]{ .886,  .937,  .855}\textbf{72.00 } & \cellcolor[rgb]{ .867,  .922,  .969}\textbf{88.33 } & \cellcolor[rgb]{ .886,  .937,  .855}\textbf{84.00 } & 69.00  & \cellcolor[rgb]{ .886,  .937,  .855}\textbf{66.00 } & \cellcolor[rgb]{ .867,  .922,  .969}\textbf{76.25 } & \cellcolor[rgb]{ .886,  .937,  .855}\textbf{73.50 } \\
    \midrule
    \href{https://huggingface.co/Qwen/Qwen-7B-Chat}{Qwen-7B-Chat} & 46.78  & 46.67  & 61.44  & 54.11  & 61.22  & 50.67  & 54.06  & 49.53  \\
    \href{https://huggingface.co/Qwen/Qwen-14B-Chat}{Qwen-14B-Chat} & 63.33  & 49.67  & 72.67  & 58.89  & 63.00  & 54.89  & 65.58  & 53.28  \\
    \href{https://huggingface.co/Qwen/Qwen-72B-Chat}{Qwen-72B-Chat} & 69.39  & 63.83  & 80.11  & 79.00  & 71.11  & 65.89  & 72.50  & 68.14  \\
    \href{https://huggingface.co/Qwen/Qwen2-7B-Instruct}{Qwen2-7B-Instruct} & 64.94  & 44.56  & 75.44  & 58.78  & 66.78  & 52.89  & 68.02  & 50.20  \\
    \href{https://huggingface.co/Qwen/Qwen2-72B-Instruct}{Qwen2-72B-Instruct} & 74.44  & \cellcolor[rgb]{ .886,  .937,  .855}\textbf{73.17$^\dag$} & 86.56  & 86.33  & 71.22  & \cellcolor[rgb]{ .886,  .937,  .855}\textbf{68.22 } & 76.67  & \cellcolor[rgb]{ .886,  .937,  .855}\textbf{75.22 } \\
    \href{https://huggingface.co/Qwen/Qwen2.5-7B-Instruct}{Qwen2.5-7B-Instruct} & 66.33  & 62.72  & 74.33  & 74.44  & 65.22  & 60.33  & 68.05  & 65.06  \\
    \href{https://huggingface.co/Qwen/Qwen2.5-14B-Instruct}{Qwen2.5-14B-Instruct} & 75.39  & 69.11  & 83.67  & 78.67  & 67.67  & 66.56  & 75.53  & 70.86  \\
    \href{https://huggingface.co/Qwen/Qwen2.5-32B-Instruct}{Qwen2.5-32B-Instruct} & 75.22  & 69.11  & \cellcolor[rgb]{ .867,  .922,  .969}\textbf{91.00 } & \cellcolor[rgb]{ .886,  .937,  .855}\textbf{86.56 } & 70.67  & 65.78  & 78.03  & 72.64  \\
    \href{https://huggingface.co/Qwen/Qwen2.5-72B-Instruct}{Qwen2.5-72B-Instruct} & \cellcolor[rgb]{ .867,  .922,  .969}\textbf{77.00$^\dag$} & 70.33  & 87.11  & 83.56  & \cellcolor[rgb]{ .867,  .922,  .969}\textbf{73.33 } & 66.67  & \cellcolor[rgb]{ .867,  .922,  .969}\textbf{78.61 } & 72.72  \\
    \midrule
    \href{https://openai.com/index/gpt-3-5-turbo-fine-tuning-and-api-updates/}{GPT-3.5-Turbo 1106} & 50.61  & 48.56  & 58.67  & 64.44  & 42.44  & 56.56  & 50.58  & 54.53  \\
    \href{https://openai.com/index/gpt-4o-mini-advancing-cost-efficient-intelligence/}{GPT-4o mini 2024-07-18} & 72.67  & 68.39  & 88.44  & 85.33  & 69.67  & 64.89  & 75.86  & 71.75  \\
    \href{https://openai.com/index/hello-gpt-4o/}{GPT-4o 2024-05-13} & \cellcolor[rgb]{ .867,  .922,  .969}\textbf{76.94 } & \cellcolor[rgb]{ .886,  .937,  .855}\textbf{73.11 } & \cellcolor[rgb]{ .867,  .922,  .969}\textbf{92.22$^\dag$} & \cellcolor[rgb]{ .886,  .937,  .855}\textbf{88.22$^\dag$} & \cellcolor[rgb]{ .867,  .922,  .969}\textbf{77.44$^\dag$} & \cellcolor[rgb]{ .886,  .937,  .855}\textbf{70.89$^\dag$} & \cellcolor[rgb]{ .867,  .922,  .969}\textbf{80.89$^\dag$} & \cellcolor[rgb]{ .886,  .937,  .855}\textbf{76.33$^\dag$} \\
    \bottomrule
    \end{tabular}}
    \caption{Evaluation Results for M\textsc{otive}B\textsc{ench} across 7 popular model families in 3 domains, including Base and CoT prompting. The best results in each series are highlighted in \textbf{Bold}, with the best overall results marked by $^\dag$.}
  \label{final_results}
\end{table*}%

\section{Experiments}
\subsection{Experimental Setup}
For each scenario, there are three types of questions presented in the form of MCQ. The scenario is only considered correct when all three questions within the same scenario are answered correctly. The specific format is shown in Appendix~\ref{question_format}.

We evaluate LLMs in two settings: one using vanilla prompting with task instructions (Base), and the other employing chain-of-thought reasoning (CoT)~\cite{wei2022chain}. The prompts we used are detailed in Appendix~\ref{sec:prompts}. Given that LLMs have been shown to exhibit a bias towards the order of choices~\cite{zheng2023large}, we introduce random variations in the choice order by generating 6 permutations. This ensures the correct option appears in all possible positions, while the incorrect options are randomly shuffled each time. We report the average of the 6 results as the final performance.

We evaluate 29 popular LLMs, as listed in Appendix~\ref{sec:benchmarkingllms}. 
For all open-source models, we use the vLLM\footnote{https://docs.vllm.ai} and set the temperature parameter to 0 to ensure result stability. For closed-source models, we access them through the Azure OpenAI API\footnote{https://azure.microsoft.com}.

\begin{figure*}[ht]
  \centering
  \includegraphics[scale=0.4]{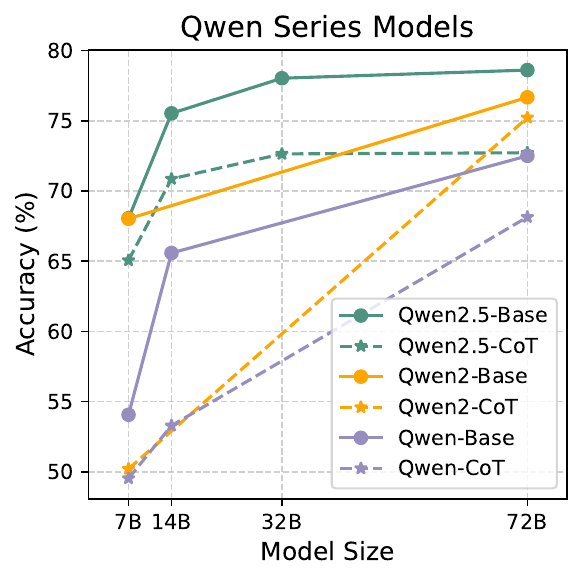}
  \hspace{-0.05in}
  \includegraphics[scale=0.4]{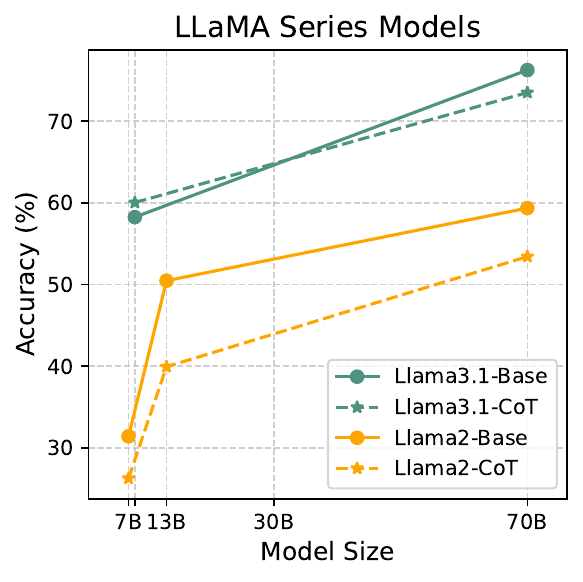}
  \hspace{-0.05in}
  \includegraphics[scale=0.4]{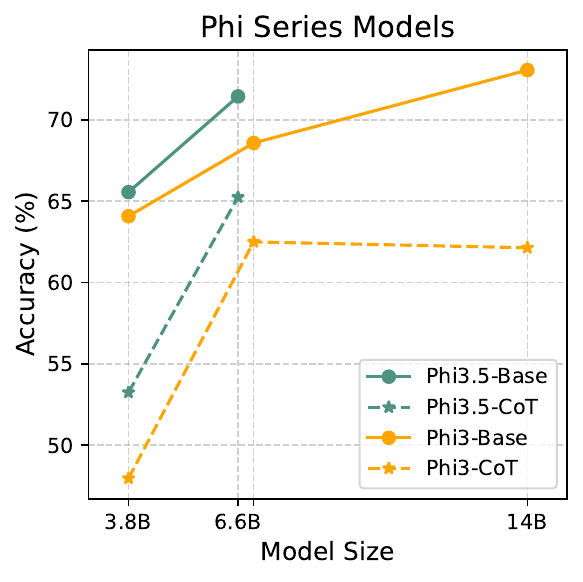}
  \hspace{-0.05in}
  \includegraphics[scale=0.4]{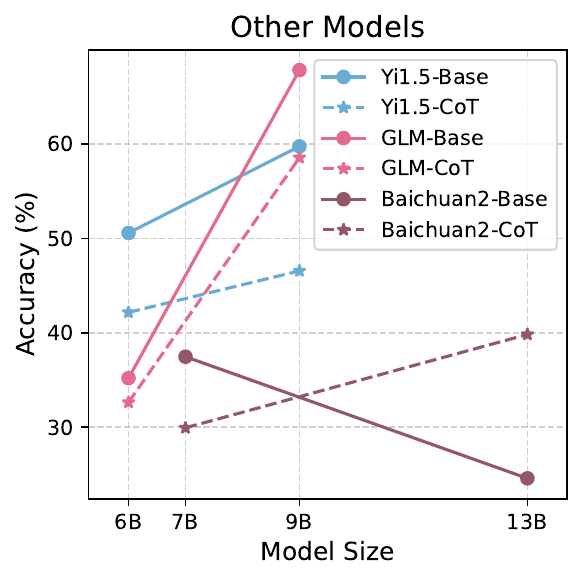}
  \caption{Scaling of various model families across their different versions and sizes in motivational reasoning ability.}
  \label{fig:perftrend}
\end{figure*}

\begin{table*}[htbp]
  \centering
  \resizebox{\textwidth}{!}{
    \begin{tabular}{l|cc|cc|cc|cc|cc|c}
    \toprule
    \multicolumn{1}{c|}{\multirow{2}[1]{*}{\textbf{Needs Hierarchy}}} & \multicolumn{2}{c|}{\textbf{GPT series}} & \multicolumn{2}{c|}{\textbf{LLaMA series}} & \multicolumn{2}{c|}{\textbf{Qwen series}} & \multicolumn{2}{c|}{\textbf{Phi series}} & \multicolumn{2}{c|}{\textbf{Yi series}} & \multirow{2}[1]{*}{\makebox[0.08\textwidth][c]{\textbf{AVG.}}} \\
          & \makebox[0.08\textwidth][c]{4o}    & \makebox[0.08\textwidth][c]{4o mini} & \makebox[0.08\textwidth][c]{3.1 70B} & \makebox[0.08\textwidth][c]{3.1 8B} & \makebox[0.08\textwidth][c]{2.5 72B} & \makebox[0.08\textwidth][c]{2.5 32B} & \makebox[0.08\textwidth][c]{3.5 MoE} & \makebox[0.08\textwidth][c]{3 Medium} & \makebox[0.08\textwidth][c]{1.5 34B} & \makebox[0.08\textwidth][c]{1.5 9B} &  \\
    \midrule
    \textbf{Level 1: Physiological} & \cellcolor[rgb]{ .988,  .549,  .231}85.83  & \cellcolor[rgb]{ .992,  .725,  .486}80.00  & \cellcolor[rgb]{ .988,  .549,  .231}84.31  & \cellcolor[rgb]{ .988,  .549,  .231}75.97  & \cellcolor[rgb]{ .992,  .871,  .745}76.67  & \cellcolor[rgb]{ .988,  .549,  .231}88.33  & \cellcolor[rgb]{ .988,  .549,  .231}80.69  & \cellcolor[rgb]{ .988,  .549,  .231}82.50  & \cellcolor[rgb]{ .992,  .725,  .486}69.86  & \cellcolor[rgb]{ .992,  .725,  .486}64.44  & \cellcolor[rgb]{ .988,  .549,  .231}78.86  \\
    \textbf{Level 2: Safety} & \cellcolor[rgb]{ .992,  .725,  .486}83.20  & \cellcolor[rgb]{ .992,  .871,  .745}79.69  & \cellcolor[rgb]{ .992,  .725,  .486}82.49  & \cellcolor[rgb]{ .992,  .871,  .745}64.31  & \cellcolor[rgb]{ .992,  .725,  .486}81.50  & \cellcolor[rgb]{ .992,  .725,  .486}86.45  & \cellcolor[rgb]{ .992,  .725,  .486}80.58  & \cellcolor[rgb]{ .992,  .871,  .745}74.34  & \cellcolor[rgb]{ .992,  .871,  .745}68.32  & \cellcolor[rgb]{ .988,  .549,  .231}68.72  & \cellcolor[rgb]{ .992,  .725,  .486}76.96  \\
    \textbf{Level 3: Love \& Belonging} & 77.27  & 72.86  & 69.93  & \cellcolor[rgb]{ .996,  .957,  .922}62.68  & 74.05  & 76.16  & \cellcolor[rgb]{ .996,  .957,  .922}64.81  & \cellcolor[rgb]{ .996,  .957,  .922}68.71  & 62.49  & 54.28  & 68.32  \\
    \textbf{Level 4: Esteem} & \cellcolor[rgb]{ .996,  .957,  .922}80.01  & \cellcolor[rgb]{ .996,  .957,  .922}77.27  & \cellcolor[rgb]{ .996,  .957,  .922}74.83  & 61.07  & \cellcolor[rgb]{ .996,  .957,  .922}74.92  & \cellcolor[rgb]{ .996,  .957,  .922}81.58  & 62.31  & 65.32  & \cellcolor[rgb]{ .996,  .957,  .922}65.96  & \cellcolor[rgb]{ .996,  .957,  .922}57.55  & \cellcolor[rgb]{ .996,  .957,  .922}70.08  \\
    \textbf{Level 5: Self-actualization} & \cellcolor[rgb]{ .992,  .871,  .745}82.83  & \cellcolor[rgb]{ .988,  .549,  .231}80.65  & \cellcolor[rgb]{ .992,  .871,  .745}81.40  & \cellcolor[rgb]{ .992,  .725,  .486}69.20  & \cellcolor[rgb]{ .988,  .549,  .231}84.52  & \cellcolor[rgb]{ .992,  .871,  .745}82.20  & \cellcolor[rgb]{ .992,  .871,  .745}74.35  & \cellcolor[rgb]{ .992,  .725,  .486}76.04  & \cellcolor[rgb]{ .988,  .549,  .231}70.57  & \cellcolor[rgb]{ .992,  .871,  .745}63.57  & \cellcolor[rgb]{ .992,  .871,  .745}76.53  \\
    \bottomrule
    \end{tabular}}
  \caption{Hierarchy of needs-oriented evaluation results for different model families and their strongest models.}
  \label{tab:Hierarchyresult}
  \vspace{-1em}
\end{table*}%

\subsection{Main Results}
Table~\ref{final_results} summarizes the performance of various LLMs across three domains, with detailed task-specific results in Appendix~\ref{sec:detailed_results}. Below, we analyze the results and highlight several key findings.


First of all, since our M\textsc{otive}B\textsc{ench} inherently reflect human consensus, the accuracy of each model serves as an indicator of its capacity for human-like motivation-behavior reasoning. It is evident that even the most advanced model, GPT-4o, exhibits a performance gap compared to human-level reasoning, with an evaluation result of 80.89\%. Furthermore, analysis of questions where GPT-4o failed reveals persistent differences in reasoning patterns with human. These discrepancies will be examined in detail in Section~\ref{analysis} Insight 2.

Secondly, among all the LLMs we evaluated, GPT-4o demonstrates the strongest capability in M\textsc{otive}B\textsc{ench}. Notably, within the open-source model series, the Qwen2.5 series demonstrates strong performance, with smaller models (14B, 32B) achieving capabilities comparable to the 72B model and even GPT-4o. Similarly, LLaMA 3.1-70B also shows good results. However, other series, especially Baichuan2, exhibit weaker reasoning capabilities for motivation and behavior tasks. From the perspective of model size, small-scale models (<10B) achieve an average accuracy of 57.16\%, medium-scale models (10B-34B) 61.83\%, and large-scale models (>34B) 71.34\%. These findings suggest a clear trend of improved motivational and behavioral reasoning ability as model size increases. The pattern is better visualized in Figure~\ref{fig:perftrend}.

Thirdly, CoT does not enhance the performance of LLMs in M\textsc{otive}B\textsc{ench}. In fact, results from most models indicate that CoT may lead to a decrease in performance. This effect is particularly pronounced in models with smaller parameter sizes (<=34B), where accuracy drops by 6.88\%, compared to a 3.14\% decrease in larger models (>=70B). This decline may occur because CoT simplifies tasks, but when the model's reasoning diverges from human cognitive patterns on motivations and behaviors, it hampers alignment with human cognition, reducing performance. In contrast, upgrading models and increasing their size significantly improve human-like motivational reasoning, as illustrated in Figure~\ref{fig:perftrend}. 

Recent literature increasingly suggests that CoT prompting does not always improve performance on certain reasoning tasks, especially those involving commonsense reasoning rather than knowledge-based or symbolic reasoning~\citep{spragueCoTNotCoT2024,liuMindYourStep2024}. Our further investigation reveals that CoT can even be detrimental when it comes to reasoning about internal human motivations and their corresponding outward behaviors. Through case analyses (Appendix~\ref{app.case}), we identified two main issues: (1) The consensus answer is sometimes the one that best fits the context, rather than the one that is most logically deduced. Step-by-step reasoning tends to prioritize maximizing logical coherence or constructing generalized motivation-behavior chains. However, humans are not fully rational agents and are more inclined to choose answers that align with situational cues and intuitive judgments. (2) There is a tendency toward overthinking, where the model excessively weighs multiple options, leading to deviation from the optimal choice. This pattern more closely resembles individuals with indecisive tendencies, whereas the consensus answer reflects the intuitive judgment of most people—who can generally identify the option that best aligns with the character’s motivations and context without extensive deliberation.

Another interesting finding is that, LLMs perform poorly in understanding \textit{"love \& belonging"} needs, which are related to emotional aspects. In Table~\ref{tab:Hierarchyresult} we break the overall score into five motivation aspects, as introduced in Section~\ref{sec.needs}. While some studies suggest that LLMs can provide emotional value comparable to or even surpassing that of humans—for instance, the Replika chatbot reduced suicidal ideation for 3 percent of users~\cite{maples2024loneliness}—they still exhibit limitations in emotional understanding and reasoning~\cite{sabour2024emobench}. This may be attributed to: (1) LLMs excel in providing emotional value by mimicking surface-level language patterns, creating a sense of understanding without deep causal reasoning. (2) The expression of \textit{"love \& belonging"} needs in the text data is often implicit or ambiguous, as it primarily involves internal processes. Since the model lacks direct exposure to comprehensive and real-world social psychology case data, it struggles to handle such issues effectively and the responses often fail to capture the emotional depth required in complex human relationships.

\subsection{In-Depth Analysis}
\label{analysis}
\paragraph{Insight 1: Comparison with Existing Benchmarks.}
We aim to introduce a new evaluation dimension—{Motive}. To study the difference between M\textsc{otive}B\textsc{ench} and existing benchmarks, we leverage   LiveBench~\cite{white2024livebench}, which typically assess general capabilities like coding, mathematics, reasoning, data analysis, language comprehension, and instruction following. 
\begin{figure}[ht]
    \centering
    \includegraphics[width=0.9\linewidth]{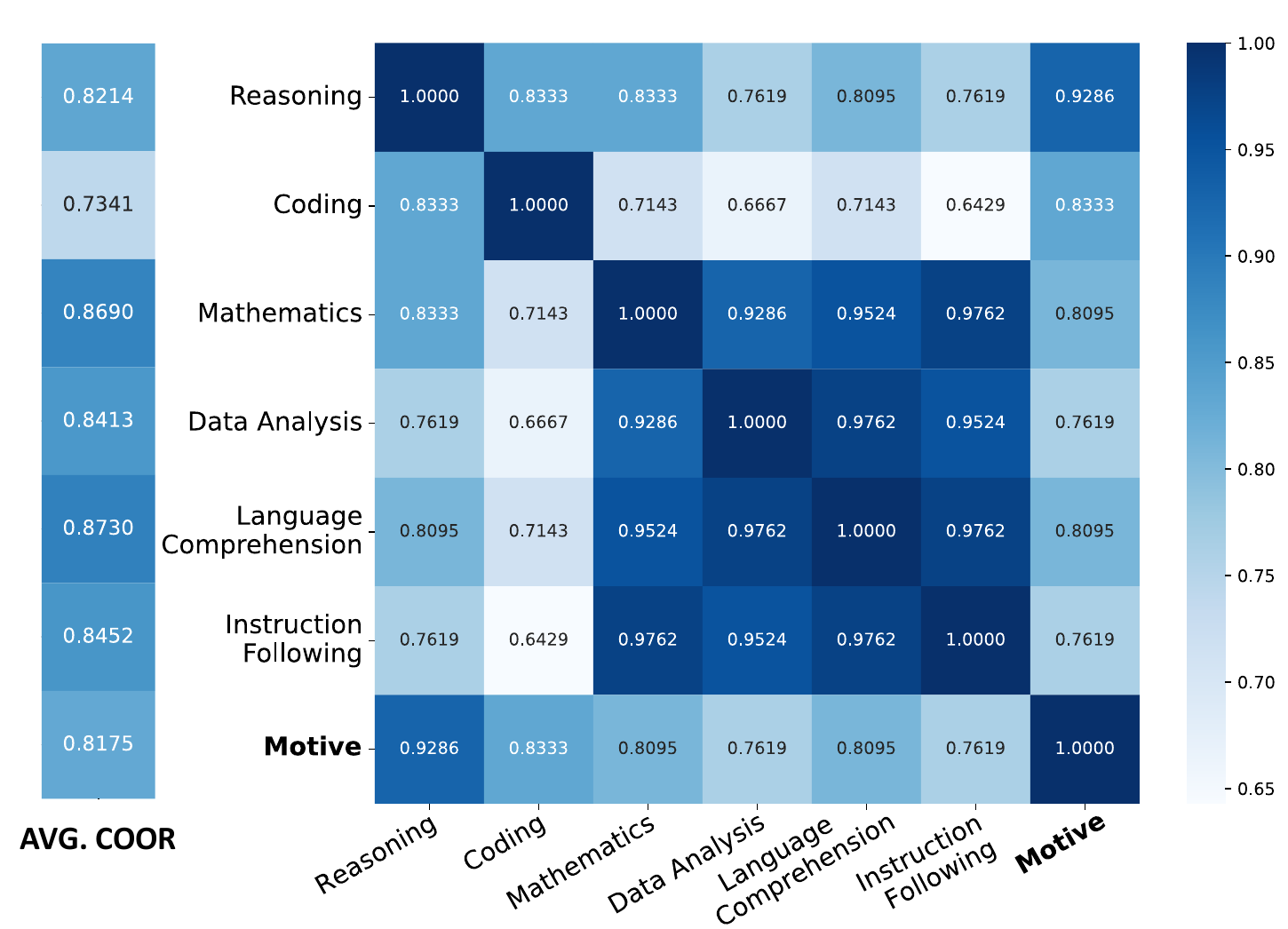}
    \caption{The correlation coefficients between the evaluation results of six general capabilities and the Motive.}
    \label{corr}
    \vspace{-1em}
\end{figure}

Figure~\ref{corr} shows the Pearson correlation coefficients~\cite{cohen2009pearson} of rankings across different ability dimensions for several popular LLMs. Notably, the Motive dimension appears robust and distinctive from other evaluation dimensions, with an average correlation coefficient of 0.8175. This suggests that by introducing the Motive dimension, we can gain new insight into underlying patterns or relationships in human capabilities that conventional evaluation metrics tend to overlook.

\begin{figure*}[t]
    \centering
    \includegraphics[width=\linewidth]{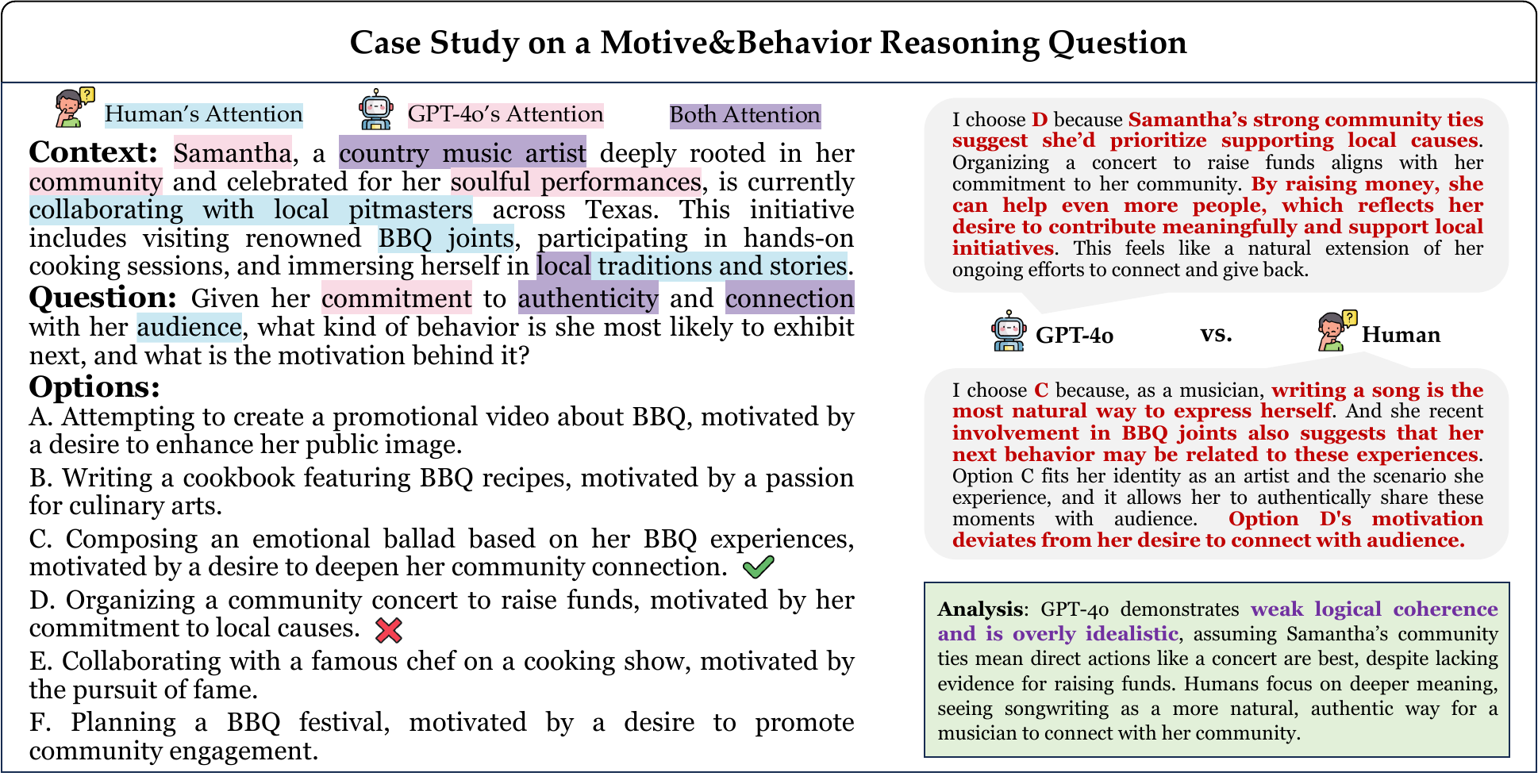}
    \caption{A case study on motivation and behavior reasoning: we analyze GPT-4o's attention within the question and its explicitly generated reasoning process to uncover key differences between its approach and human cognition.}
    \label{case_study}
\end{figure*}

\paragraph{Insight 2: Differences Between GPT-4o and Human in Motivation and Behavior.} In Table~\ref{final_results}, we observe that GPT-4o is the most advanced model in our benchmark. Therefore, we are curious to investigate the situations in which this model fails to demonstrate human-like intelligence. For questions answered incorrectly by GPT-4o, we examine its reasoning and thought processes. We have summarized the following findings:

\textbf{1) Over-Rationalization, Lacking Emotional Insight.} GPT-4o often relies on logical reasoning, neglecting broader practical experience or emotional context, leading to reasoning that may be disconnected from real-world complexities.

\textbf{2)Weak Logical Precision, Prone to General Assumptions.} GPT-4o's reasoning can be overly simplistic, often based on general assumptions or external knowledge, without fully addressing the specific context or details of a situation.

\textbf{3)Overly Idealistic, Ignoring Complex Realities.} It tends to assume people follow social norms or moral codes, ignoring more complex or real-world challenges that could affect behavior.

\textbf{4)Lack of Awareness of Behavioral Impact.} It may prioritize actions that seem easy or plausible but fail to consider their actual effectiveness or long-term impact in real-life scenarios.

Figure~\ref{case_study} presents an example of GPT-4o demonstrating weak logical coherence and overly idealistic. Detailed discussion and interpretation, and more cases can be found in the Appendix~\ref{app:cases}.

\paragraph{Insight 3: Limitations of LLMs in Complete Data Annotation.}
Recently, many researchers have been curious about whether "large language models replace human annotators" or "replace human participants in social science experiments". By using our M\textsc{otive}B\textsc{ench} as the lens, we find that LLMs for annotation presents several challenges:

\textbf{1) Logical or Factual Errors.} LLMs may generate inaccurate or misleading questions due to limited understanding of psychological theories, resulting in the content of the questions or options lacking logical consistency. Therefore, we introduce a manual correction step to ensure the reliability of the reasoning questions.

\textbf{2) Limited Understanding of Human Dynamics.} The tasks we consider often involve nuanced psychological and sociological dimensions, which LLMs may struggle to accurately capture due to the complexity of human thought processes.

\textbf{3)Annotator-specific Bias.} Relying on a single LLM introduces biases, distorting the objective interpretation of psychological data. 
To minimize such biases, our framework uses multiple models at different stages of the pipeline and incorporates human labeling to revise questions at the final step.

\paragraph{Insight 4: Trends of Motivational Intelligence in LLMs of Varying Sizes}Our experiments with LLMs of varying sizes reveal distinct trends in motivational intelligence. Smaller models (6B-13B) struggle with complex reasoning, long-chain inferences, and emotional understanding in the informative contexts we provide, often producing superficial responses. Medium-sized models (13B-34B) show improved performance, are able to capture motivational cues, provided better contextual understanding and somewhat human-like thinking patterns. In addition, large models (70B+), particularly GPT-4o and Qwen2.5-72B, excel in complex reasoning and nuanced motivational modeling. They integrate scenario and character's profile information effectively, understanding motivations, emotions, and logical relationships, thus showing behavior patterns closer to human intelligence.

However, current LLMs still differ from humans in some aspects of motivational reasoning (see Insight 2), posing challenges in fully open-ended, dynamic scenarios like real-time strategy games and unpredictable social interactions—key frontiers for agent simulation and decision-making systems.



\section{Related Work}
Currently, the intellectual capabilities of LLMs have reached an unexpected level, but they also display a certain unreliability—sometimes becoming confused, while at other times demonstrating abilities that far exceed human in specific evaluations. But how far are they from human-like cognition? An increasing number of studies are now approaching this issue from Theory of Mind (ToM) perspective. \citet{sap2022neural} find that GPT-3 lacks social intelligence by using S\textsc{ocial}IQ\textsc{a}~\cite{sap2019socialiqa} and ToMi~\cite{le2019revisiting}. \citet{ullman2023large} demonstrate that small variations that preserve the principles of ToM can drastically alter the outcomes. Similar results can also be found from \citet{shapira2023clever}. \citet{strachan2024testing} observe that GPT-4 excelled at identifying indirect requests, false beliefs, and misdirection, but struggled with detecting faux pas, and it still lags behind humans in overall ToM performance~\cite{chen2024tombench}.
In addition, some studies have conducted experiments from a behavioral perspective. \citet{xiao2023far} reveal that current LLMs struggle to align their behaviors with assigned characters. Furthermore, \citet{zhou2023sotopia} and \citet{wang2024towards} propose interactive and sandbox benchmarks, showing GPT-4 excels in conversational scenarios but struggles to exhibit social commonsense reasoning and deal with social tasks~\cite{lee2024towards}.

Different from existing research, we aim to explore a new dimension—Motive, which examines the alignment between LLMs and human behavior in dynamic and unstructured environments. 

\section{Conclusions}
We introduce M\textsc{otive}B\textsc{ench}, the first systematic benchmark to evaluate the human-like motivational and behavioral reasoning ability of LLMs with detailed, realistic situations. 
Our results reveal that even the most advanced LLMs still fall short in achieving human-like motivational reasoning. Furthermore, most LLMs struggle with understanding \textit{"love \& belonging"} needs. In-depth analysis reveals specific differences in motivational reasoning between current LLMs and humans. By introducing M\textsc{otive}B\textsc{ench}, we aim to provide insights from a new dimension, enabling future models to exhibit more human-like cognition processes.

\section*{Limitations}
\subsection*{Fully Automated Questions Generation}
In the current pipeline for generating questions in M\textsc{otive}B\textsc{ench}, we still rely on human effort to manually check and revise the quality of questions. This approach poses challenges for automatically refreshing the benchmark to avoid data contamination for future LLM releases and for scaling to a larger set of test questions.

To address these challenges, it is necessary to train a revision model using our existing manually corrected data, with the goal of fully automating the entire question generation pipeline. This would enable the benchmark to dynamically update itself, thereby maximizing its value in the rapidly evolving era of LLMs. To achieve this, a potential approach involves leveraging advanced techniques, focusing particularly on fine-tuning existing pre-trained models with our manually corrected dataset. This process will help the model learn the intricate patterns and nuances required for high-quality question generation. Additionally, we plan to incorporate continuous learning mechanisms, allowing the model to adapt to new data and evolving trends. By doing so, we aim to enhance not only the accuracy and relevance of the generated questions but also ensure that the benchmark remains aligned with the latest developments in the field of large language models. The result will be a more dynamic, scalable, and efficient system capable of keeping pace with advancements in AI technology.
\subsection*{From Situational QA to Realistic Simulations}
M\textsc{otive}B\textsc{ench} fundamentally employs the "situational question-answering" paradigm, where LLMs are prompted to answer questions about the next immediate step in various scenarios. However, this paradigm still deviates from real-world human social activities, where individuals take a sequence of actions, form longer life stories, and behave spontaneously without being asked, "what will you do?". 

To overcome this limitation, we can consider using the paradigm of LLMs performing role-playing in a simulation sandbox~\cite{wang2024characterbox}. Given an initial scenario, LLMs act as specific characters with preset personas to engage in daily activities or achieve goal-oriented tasks. By analyzing the behavior trajectories elicited in the sandbox, we can assess LLMs' motivational reasoning and proactive action-taking capabilities in a more comprehensive manner.

\section*{Acknowledgements}
We thank the reviewers for their valuable feedback. This work was supported by the Public Computing Cloud at Renmin University of China and the Fund for Building World-Class Universities (Disciplines) at Renmin University of China.

\bibliography{custom}

\appendix
\section{Details of the Hierarchy of Needs}
\begin{figure*}
    \centering
    \includegraphics[width=0.98\linewidth]{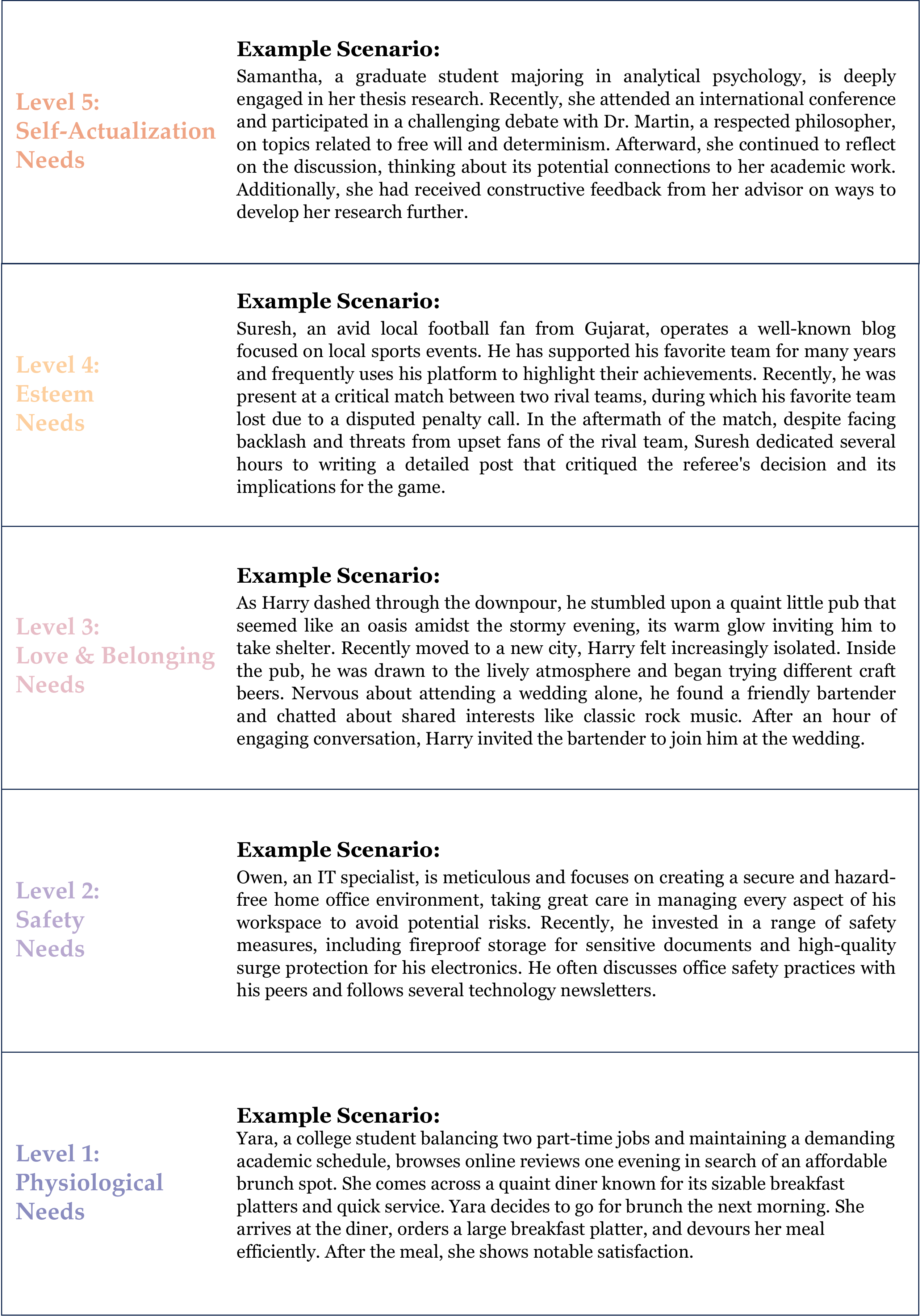}
    \caption{Example scenarios for Maslow's hierarchy of needs in M\textsc{otive}B\textsc{ench}.}
    \label{scenarios}
\end{figure*}

Maslow's hierarchy of needs~\cite{maslow1943theory} is a motivational theory proposed by American psychologist Abraham Maslow in 1943, which explains the prioritization and fulfillment of human needs. He categorized human needs into five levels, ranging from basic survival needs to higher psychological needs, ultimately culminating in self-actualization. Below is a detailed description of each level of need, along with corresponding example scenarios from our M\textsc{otive}B\textsc{ench} as shown in Figure~\ref{scenarios}.
\begin{itemize}
    \item \textbf{Physiological Needs}: Physiological needs are the most fundamental survival needs for humans, including air, water, food, warmth, and sleep. These needs form the base of the hierarchy, and only once they are met can individuals engage in other activities.
    \item \textbf{Safety Needs}: Once physiological needs are met, humans seek security, which includes physical safety, financial stability, health, and environmental consistency. This need reflects an individual's desire for order, protection, and predictability in the future.
    \item \textbf{Love and Belonging Needs}: Also known as social needs, people seek to build interpersonal relationships, gain friendships, experience love, and feel a sense of belonging to a social group. This need involves the desire to integrate into society, be accepted, and interact with others.
    \item \textbf{Esteem Needs}: Esteem needs are divided into intrinsic esteem (self-confidence, self-respect, and independence) and extrinsic esteem (recognition, appreciation, and status from others). Satisfying these needs boosts an individual's sense of self-worth and social standing.
    \item \textbf{Self-Actualization Needs}: The highest level of need, self-actualization refers to the individual's pursuit of fulfilling their potential and achieving their ideal self. Self-actualization is typically manifested through creativity, personal growth, and realizing one's own value.
\end{itemize}

Furthermore, the Reiss Motivation Profile (RMP)~\cite{reiss2004multifaceted}, proposed by psychologist Steven Reiss, is also a theoretical model that aims to understand an individual's motivation by analyzing their preferences across 16 fundamental needs. Each person responds differently to these 16 needs, and this variation determines their behavior, decision-making, and lifestyle. The core assumption of the RMP model is that individual behaviors are driven by these needs, and the intensity of these needs shapes one's behavioral patterns.

Fundamental human needs such as \textbf{Curiosity}, \textbf{Idealism}, and \textbf{Honor} motivate individuals to explore knowledge, uphold ethical principles, and seek validation from society, representing a pursuit of cognitive growth and moral fulfillment. Desires for control and autonomy are embodied in \textbf{Independence} and \textbf{Power}, where those driven by independence prioritize freedom and self-governance, while those seeking power focus on exerting influence and leadership. After satisfying cognitive and control-oriented needs, social and emotional desires come to the forefront: \textbf{Acceptance}, \textbf{Family}, and \textbf{Romance} emphasize the importance of close relationships and emotional intimacy, whereas \textbf{Social Contact} and \textbf{Status} reflect the longing for social integration and external recognition. Practical concerns for stability are seen in the needs for \textbf{Order} and \textbf{Saving}, which focus on creating structure and security in life. Simultaneously, the pursuit of personal well-being emerges through \textbf{Tranquility} and \textbf{Physical Activity}, highlighting the desire for peace of mind and physical health. Finally, more primal or visceral needs, such as \textbf{Vengeance} and \textbf{Eating}, speak to coping with perceived injustice and savoring the pleasures of sensory experiences, offering a reminder of the complexities inherent in human motivation.

\section{Question Format Example}
\label{question_format}
\begin{figure}
    \centering
    \includegraphics[width=\linewidth]{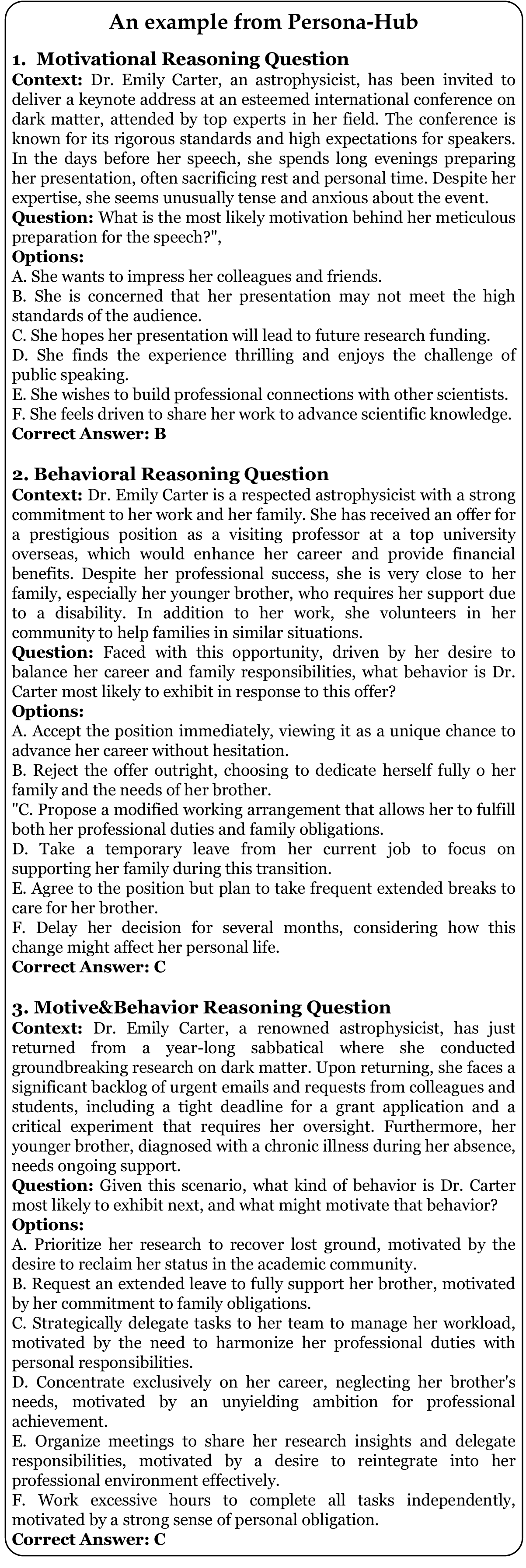}
    \caption{An example question from Persona-Hub.}
    \label{fig:qa_example}
\end{figure}

Figure~\ref{fig:qa_example} shows an example of three types of tasks in the same scenario, from Persona-Hub.

\section{Human Annotation Results}
\label{annatation}
\subsection{Recruitment of Data Annotators}
From Dec.18, 2024 to Jan.13, 2025, we recruited participants through a crowdsourcing platform to annotate questions. A total of 28 individuals registered for the study. Based on their performance in a small set of trial annotations, 15 participants were selected (rejection rate: 46.43\%).

To comprehensively assess the impact of demographics and perspectives on answer selection, we recruited annotators from \textbf{diverse backgrounds}, including fields such as minority language translation, public administration, metallurgical engineering, law, chemistry, IoT engineering, art, social demography, and computer science. Their \textbf{education levels} range from undergraduate to PhD, with an \textbf{average age} of 23.7.

\subsection{Requirements for Annotation Task}
Each annotator is responsible for evaluating both the \textbf{correctness} and \textbf{reasonableness} of the multiple-choice questions. The task is divided into two main components:
\begin{itemize}
    \item For each MCQ, annotators assign a total of 6 points across the available options based on their assessment of the correctness of each option. Higher scores indicate greater correctness. The points must be allocated as whole integers, with the total sum of points for each question always equaling 6.
    \item Annotators also evaluate the overall reasonableness of each question. A question is deemed unreasonable if any of the following conditions are met: 1) The scenario presented is logically inconsistent or flawed; 2) Insufficient background information is provided, making it impossible to derive any meaningful answer; 3) None of the answer options are reasonable or relevant; 4) There are three or more answer options with equally high correctness scores, causing ambiguity and reducing the clarity of the question.
\end{itemize}

To ensure objectivity and thoroughness, each question is independently reviewed by five different annotators. Each annotator dedicates 8 hours to annotate a set of 200 questions. To maintain the integrity of the task and prevent any potential answer leakage or bias, annotators are restricted to viewing and annotating only one question per scenario. This approach ensures that they have no exposure to related questions within the same context, safeguarding the independence of their judgments.

\subsection{Analysis of Annotation Results}
A comprehensive analysis of annotation quality, question consistency, and response accuracy reveals the following key findings:
\begin{itemize}
    \item \textbf{Strong inter-annotator consistency}: Pearson correlation coefficients between annotator pairs range from 0.692 to 0.868, indicating a high level of alignment in their evaluations. Moreover, the absence of unreasonable cases highlights the clarity and robustness of the questions, as well as the accuracy of the annotations.
    \item \textbf{High agreement rates}: 98.17\% of the questions achieved agreement among at least three annotators, while 82.50\% reached consensus among four annotators. This demonstrates exceptional consistency in judgment and further confirms the reasonableness of the questions.
    \item \textbf{Accurate consensus answers}: By aggregating the weights assigned by annotators to each option and selecting the one with the highest cumulative score, we find that consensus answers match our pre-defined correct answers at a rate of 93.00\%, thereby validating the accuracy of our predefined answers.
\end{itemize}
These annotation results suggest that the questions were generally perceived as well-constructed, reasonable, and logically sound by the annotators. The high consistency in the annotations further indicates that the evaluative criteria were clear and effectively applied across different annotators. Additionally, the analysis reveals that demographic factors, such as individual backgrounds or perspectives, appear to have little to no influence on the final answer choices, suggesting that our M\textsc{otive}B\textsc{ench} can make an objective and reasonable assessment of motivational intelligence.

\begin{table}[htbp]
  \centering
  \caption{Overview of Annotation Results}
  \resizebox{0.47\textwidth}{!}{
    \begin{tabular}{l|c}
    \toprule
    \makebox[0.35\textwidth][l]{\textbf{Number of Incorrect Questions}} & \makebox[0.1\textwidth][c]{0} \\
    \midrule
    \makebox[0.35\textwidth][l]{\textbf{Number of Correct Questions}} & \makebox[0.1\textwidth][c]{600} \\
    \midrule
    \textbf{$\geq$ 3 Agreement (\%)} & 98.17  \\
    \midrule
    \textbf{$\geq$ 4 Agreement (\%)} & 82.50  \\
    \midrule
    \textbf{Agreement w/ Answers (\%)} & 93.00  \\
    \bottomrule
    \end{tabular}}
  \label{tab:annotation}%
\end{table}%

\section{Evaluated LLMs}
\label{sec:benchmarkingllms}
We evaluate 29 popular LLMs across a range of parameter sizes, including several models from the GPT series~\cite{hurst2024gpt} (GPT-4o 2024-05-13, GPT-4o mini 2024-07-18, and GPT-3.5-Turbo 1106), the LLaMA series~\cite{meta2024introducing,touvron2023llama} (LLaMA 3.1 and LLaMA 2), the Qwen series~\cite{hui2024qwen2,yang2024qwen2,bai2023qwen} (Qwen 2.5, Qwen 2, and Qwen), the Phi series~\cite{abdin2024phi} (Phi 3.5 and Phi 3), the GLM series~\cite{glm2024chatglm} (ChatGLM 3 and GLM 4), as well as other models like Baichuan 2~\cite{yang2023baichuan} and Yi 1.5~\cite{young2024yi}. These models span a wide spectrum of architectures and parameter configurations, offering a comprehensive evaluation of current LLM performance across various tasks and benchmarks. All of our experiments are conducted on a machine with four A100 80GB GPUs.

\begin{table*}[htbp]
  \centering
  \resizebox{\textwidth}{!}{
    \begin{tabular}{l|ccc|ccc|ccc}
    \toprule
\multicolumn{10}{l}{$\mathbb{MRQ}$: Motivational Reasoning Question} \\
    \multicolumn{10}{l}{$\mathbb{BRQ}$: Behavioral Reasoning Question} \\
    \multicolumn{10}{l}{$\mathbb{MBQ}$: Motive\&Behavior Reasoning Question} \\
    \midrule
    \multicolumn{1}{c|}{\multirow{2}[2]{*}{\textbf{Base Method}}} & \multicolumn{3}{c|}{\textbf{Vitural Profiles}} & \multicolumn{6}{c}{\textbf{Real-life Scenarios}} \\
          & \multicolumn{3}{c|}{\textbf{Persona-Hub}} & \multicolumn{3}{c}{\textbf{Amazon}} & \multicolumn{3}{c}{\textbf{Blogger}} \\
    \midrule
    \textbf{LLMs} & \textbf{$\mathbb{MRQ}$} & \textbf{$\mathbb{BRQ}$} & \textbf{$\mathbb{MBQ}$} & \textbf{$\mathbb{MRQ}$} & \textbf{$\mathbb{BRQ}$} & \textbf{$\mathbb{MBQ}$} & \textbf{$\mathbb{MRQ}$} & \textbf{$\mathbb{BRQ}$} & \textbf{$\mathbb{MBQ}$} \\
    \midrule
    Baichuan2-7B-Chat & \cellcolor[rgb]{ 1,  .949,  .8}\textbf{79.00 } & 66.56  & \cellcolor[rgb]{ 1,  .918,  .898}\textbf{61.83 } & 69.44  & 68.89  & \cellcolor[rgb]{ 1,  .918,  .898}\textbf{74.22 } & \cellcolor[rgb]{ 1,  .949,  .8}\textbf{81.11 } & \cellcolor[rgb]{ .91,  .898,  .953}\textbf{64.00 } & \cellcolor[rgb]{ 1,  .918,  .898}\textbf{69.33 } \\
    Baichuan2-13B-Chat & 69.06  & \cellcolor[rgb]{ .91,  .898,  .953}\textbf{67.67 } & 41.28  & \cellcolor[rgb]{ 1,  .949,  .8}\textbf{72.33 } & \cellcolor[rgb]{ .91,  .898,  .953}\textbf{72.67 } & 63.00  & 48.33  & 61.78  & 48.89  \\
    \midrule
    ChatGLM3-6B & 65.72  & 64.06  & 58.61  & 59.56  & 64.22  & 59.22  & 78.33  & 70.11  & 75.44  \\
    GLM4-9B-Chat & \cellcolor[rgb]{ 1,  .949,  .8}\textbf{91.06 } & \cellcolor[rgb]{ .91,  .898,  .953}\textbf{84.22 } & \cellcolor[rgb]{ 1,  .918,  .898}\textbf{80.33 } & \cellcolor[rgb]{ 1,  .949,  .8}\textbf{94.67 } & \cellcolor[rgb]{ .91,  .898,  .953}\textbf{88.22 } & \cellcolor[rgb]{ 1,  .918,  .898}\textbf{92.11 } & \cellcolor[rgb]{ 1,  .949,  .8}\textbf{92.78 } & \cellcolor[rgb]{ .91,  .898,  .953}\textbf{78.33 } & \cellcolor[rgb]{ 1,  .918,  .898}\textbf{86.67 } \\
    \midrule
    Yi1.5-6B-Chat & 87.33  & 78.50  & 67.11  & 83.00  & 83.78  & 79.11  & 88.33  & 72.22  & 73.33  \\
    Yi1.5-9B-Chat & 88.17  & \cellcolor[rgb]{ .91,  .898,  .953}\textbf{83.22 } & 77.28  & 91.33  & 84.22  & 85.11  & 85.78  & 72.00  & \cellcolor[rgb]{ 1,  .918,  .898}\textbf{82.22 } \\
    Yi1.5-34b-Chat & \cellcolor[rgb]{ 1,  .949,  .8}\textbf{92.33 } & 81.89  & \cellcolor[rgb]{ 1,  .918,  .898}\textbf{79.50 } & \cellcolor[rgb]{ 1,  .949,  .8}\textbf{91.78 } & \cellcolor[rgb]{ .91,  .898,  .953}\textbf{89.78 } & \cellcolor[rgb]{ 1,  .918,  .898}\textbf{93.22 } & \cellcolor[rgb]{ 1,  .949,  .8}\textbf{91.11 } & \cellcolor[rgb]{ .91,  .898,  .953}\textbf{82.00 } & 80.78  \\
    \midrule
    Phi3-mini-4k-Instruct & 87.61  & 82.89  & 78.50  & 88.33  & 85.44  & 90.89  & 90.56  & 81.78  & 85.67  \\
    Phi3-small-8k-Instruct & 90.78  & 86.17  & 82.56  & \cellcolor[rgb]{ 1,  .949,  .8}\textbf{94.89 } & 89.89  & 94.00  & 90.22  & 79.89  & 85.00  \\
    Phi3-medium-4k-Instruct & \cellcolor[rgb]{ 1,  .949,  .8}\textbf{93.22 } & \cellcolor[rgb]{ .91,  .898,  .953}\textbf{88.78 } & 82.94  & 93.89  & \cellcolor[rgb]{ .91,  .898,  .953}\textbf{93.00 } & \cellcolor[rgb]{ 1,  .918,  .898}\textbf{95.00 } & 90.56  & \cellcolor[rgb]{ .91,  .898,  .953}\textbf{84.67 } & \cellcolor[rgb]{ 1,  .918,  .898}\textbf{89.89 } \\
    Phi3.5-mini-Instruct & 88.11  & 82.83  & 78.83  & 89.56  & 87.22  & 89.89  & 91.56  & 80.00  & 87.00  \\
    Phi3.5-MoE-Instruct & 93.06  & 88.44  & \cellcolor[rgb]{ 1,  .918,  .898}\textbf{84.56 } & 94.33  & 86.44  & 90.89  & \cellcolor[rgb]{ 1,  .949,  .8}\textbf{92.00 } & 82.78  & 88.33  \\
    \midrule
    Llama2-7B-Chat & 70.33  & 60.11  & 50.94  & 68.67  & 63.44  & 69.56  & 75.78  & 62.44  & 67.78  \\
    Llama2-13B-Chat & 82.78  & 74.22  & 69.00  & 78.56  & 79.33  & 80.44  & 90.11  & 74.22  & 80.00  \\
    Llama2-70B-Chat & 90.28  & 79.50  & 72.61  & 89.00  & 82.67  & 82.78  & 90.44  & 81.67  & 83.22  \\
    Llama3.1-8B-Instruct & 86.39  & 78.67  & 72.94  & 84.33  & 79.67  & 85.00  & \cellcolor[rgb]{ 1,  .949,  .8}\textbf{94.00 } & 82.89  & 84.78  \\
    Llama3.1-70B-Instruct & \cellcolor[rgb]{ 1,  .949,  .8}\textbf{94.72 } & \cellcolor[rgb]{ .91,  .898,  .953}\textbf{91.89 } & \cellcolor[rgb]{ 1,  .918,  .898}\textbf{85.72 } & \cellcolor[rgb]{ 1,  .949,  .8}\textbf{93.78 } & \cellcolor[rgb]{ .91,  .898,  .953}\textbf{95.56 } & \cellcolor[rgb]{ 1,  .918,  .898}\textbf{97.00 } & 87.44  & \cellcolor[rgb]{ .91,  .898,  .953}\textbf{84.78 } & \cellcolor[rgb]{ 1,  .918,  .898}\textbf{91.89 } \\
    \midrule
    Qwen-7B-Chat & 84.72  & 76.61  & 68.33  & 82.67  & 83.11  & 88.00  & 87.44  & 73.22  & 79.11  \\
    Qwen-14B-Chat & 90.94  & 86.00  & 78.89  & 89.78  & 86.67  & 92.22  & 87.44  & 74.22  & 80.00  \\
    Qwen-72B-Chat & 93.44  & 86.17  & 85.61  & 92.78  & 90.67  & 93.67  & 92.78  & \cellcolor[rgb]{ .91,  .898,  .953}\textbf{85.33 } & 89.33  \\
    Qwen2-7B-Instruct & 90.28  & 86.33  & 82.78  & 91.11  & 90.00  & 90.33  & 91.78  & 85.00  & 82.78  \\
    Qwen2-72B-Instruct & 94.50  & 89.44  & 88.39  & 95.11  & 93.00  & 95.89  & 94.44  & 83.78  & 86.89  \\
    Qwen2.5-7B-Instruct & 88.67  & 86.94  & 83.94  & 90.89  & 87.67  & 91.44  & 90.44  & 81.67  & 83.22  \\
    Qwen2.5-14B-Instruct & 94.83  & \cellcolor[rgb]{ .91,  .898,  .953}\textbf{90.50 } & 87.72  & 94.33  & 91.00  & 96.11  & 91.67  & 84.78  & 86.67  \\
    Qwen2.5-32B-Instruct & 95.22  & 90.44  & 87.44  & \cellcolor[rgb]{ 1,  .949,  .8}\textbf{97.44 } & \cellcolor[rgb]{ .91,  .898,  .953}\textbf{93.89 } & \cellcolor[rgb]{ 1,  .918,  .898}\textbf{96.56 } & 93.33  & 83.78  & 85.33  \\
    Qwen2.5-72B-Instruct & \cellcolor[rgb]{ 1,  .949,  .8}\textbf{95.39 } & 90.06  & \cellcolor[rgb]{ 1,  .918,  .898}\textbf{89.72 } & 96.11  & 93.44  & 95.56  & \cellcolor[rgb]{ 1,  .949,  .8}\textbf{94.89 } & 84.67  & \cellcolor[rgb]{ 1,  .918,  .898}\textbf{90.11 } \\
    \midrule
    GPT-3.5 & 84.00  & 79.72  & 72.94  & 86.22  & 81.33  & 81.89  & 74.89  & 61.89  & 77.89  \\
    GPT-4o mini & 92.78  & 89.61  & 87.94  & 94.78  & 94.67  & 96.89  & 90.89  & 81.56  & 86.89  \\
    GPT-4o & \cellcolor[rgb]{ 1,  .949,  .8}\textbf{93.81 } & \cellcolor[rgb]{ .91,  .898,  .953}\textbf{91.31 } & \cellcolor[rgb]{ 1,  .918,  .898}\textbf{90.06 } & \cellcolor[rgb]{ 1,  .949,  .8}\textbf{96.67 } & \cellcolor[rgb]{ .91,  .898,  .953}\textbf{95.56 } & \cellcolor[rgb]{ 1,  .918,  .898}\textbf{98.78 } & \cellcolor[rgb]{ 1,  .949,  .8}\textbf{93.89 } & \cellcolor[rgb]{ .91,  .898,  .953}\textbf{88.78 } & \cellcolor[rgb]{ 1,  .918,  .898}\textbf{92.56 } \\
    \bottomrule
    \end{tabular}}
\caption{The experimental results of the vanilla prompt-based method on the three types of tasks.}
  \label{base_results}%
\end{table*}%

\begin{table*}[htbp]
  \centering
  \resizebox{\textwidth}{!}{
    \begin{tabular}{l|ccc|ccc|ccc}
    \toprule
    \multicolumn{10}{l}{$\mathbb{MRQ}$: Motivational Reasoning Question} \\
    \multicolumn{10}{l}{$\mathbb{BRQ}$: Behavioral Reasoning Question} \\
    \multicolumn{10}{l}{$\mathbb{MBQ}$: Motive\&Behavior Reasoning Question} \\
    \midrule
    \multicolumn{1}{c|}{\multirow{2}[2]{*}{\textbf{CoT Method}}} & \multicolumn{3}{c|}{\textbf{Vitural Profiles}} & \multicolumn{6}{c}{\textbf{Real-life Scenarios}} \\
          & \multicolumn{3}{c|}{\textbf{Persona-Hub}} & \multicolumn{3}{c}{\textbf{Amazon}} & \multicolumn{3}{c}{\textbf{Blogger}} \\
    \midrule
    \textbf{LLMs} & \textbf{$\mathbb{MRQ}$} & \textbf{$\mathbb{BRQ}$} & \textbf{$\mathbb{MBQ}$} & \textbf{$\mathbb{MRQ}$} & \textbf{$\mathbb{BRQ}$} & \textbf{$\mathbb{MBQ}$} & \textbf{$\mathbb{MRQ}$} & \textbf{$\mathbb{BRQ}$} & \textbf{$\mathbb{MBQ}$} \\
    \midrule
    Baichuan2-7B-Chat & 73.39  & 61.44  & 58.00  & 61.44  & 67.44  & 71.11  & 78.11  & 58.78  & \cellcolor[rgb]{ 1,  .918,  .898}\textbf{70.00 } \\
    Baichuan2-13B-Chat & \cellcolor[rgb]{ 1,  .949,  .8}\textbf{77.06 } & \cellcolor[rgb]{ .91,  .898,  .953}\textbf{75.00 } & \cellcolor[rgb]{ 1,  .918,  .898}\textbf{64.11 } & \cellcolor[rgb]{ 1,  .949,  .8}\textbf{80.22 } & \cellcolor[rgb]{ .91,  .898,  .953}\textbf{78.44 } & \cellcolor[rgb]{ 1,  .918,  .898}\textbf{75.89 } & \cellcolor[rgb]{ 1,  .949,  .8}\textbf{79.22 } & \cellcolor[rgb]{ .91,  .898,  .953}\textbf{63.56 } & 69.89  \\
    \midrule
    ChatGLM3-6B & 73.67  & 63.56  & 54.22  & 71.78  & 71.33  & 71.89  & 72.67  & 61.78  & 70.44  \\
    GLM4-9B-Chat & \cellcolor[rgb]{ 1,  .949,  .8}\textbf{82.50 } & \cellcolor[rgb]{ .91,  .898,  .953}\textbf{81.94 } & \cellcolor[rgb]{ 1,  .918,  .898}\textbf{75.28 } & \cellcolor[rgb]{ 1,  .949,  .8}\textbf{91.67 } & \cellcolor[rgb]{ .91,  .898,  .953}\textbf{89.44 } & \cellcolor[rgb]{ 1,  .918,  .898}\textbf{90.67 } & \cellcolor[rgb]{ 1,  .949,  .8}\textbf{74.33 } & \cellcolor[rgb]{ .91,  .898,  .953}\textbf{74.00 } & \cellcolor[rgb]{ 1,  .918,  .898}\textbf{79.11 } \\
    \midrule
    Yi1.5-6B-Chat & 76.72  & 73.39  & 64.06  & 75.33  & 83.44  & 80.44  & 82.89  & 64.00  & 72.67  \\
    Yi1.5-9B-Chat & 83.94  & 69.22  & 66.72  & 83.67  & 74.00  & 79.67  & 86.11  & 72.89  & \cellcolor[rgb]{ 1,  .918,  .898}\textbf{77.33 } \\
    Yi1.5-34b-Chat & \cellcolor[rgb]{ 1,  .949,  .8}\textbf{89.44 } & \cellcolor[rgb]{ .91,  .898,  .953}\textbf{81.39 } & \cellcolor[rgb]{ 1,  .918,  .898}\textbf{77.67 } & \cellcolor[rgb]{ 1,  .949,  .8}\textbf{88.33 } & \cellcolor[rgb]{ .91,  .898,  .953}\textbf{87.89 } & \cellcolor[rgb]{ 1,  .918,  .898}\textbf{89.56 } & \cellcolor[rgb]{ 1,  .949,  .8}\textbf{87.00 } & \cellcolor[rgb]{ .91,  .898,  .953}\textbf{75.67 } & 76.33  \\
    \midrule
    Phi3-mini-4k-Instruct & 81.50  & 76.67  & 69.67  & 84.33  & 84.11  & 81.44  & 78.33  & 61.11  & 71.44  \\
    Phi3-small-8k-Instruct & 89.22  & 79.94  & 77.11  & \cellcolor[rgb]{ 1,  .949,  .8}\textbf{93.44 } & \cellcolor[rgb]{ .91,  .898,  .953}\textbf{90.33 } & \cellcolor[rgb]{ 1,  .918,  .898}\textbf{92.11 } & 89.67  & \cellcolor[rgb]{ .91,  .898,  .953}\textbf{76.78 } & \cellcolor[rgb]{ 1,  .918,  .898}\textbf{80.67 } \\
    Phi3-medium-4k-Instruct & 89.17  & 79.72  & 80.39  & \cellcolor[rgb]{ 1,  .949,  .8}\textbf{93.44 } & 85.11  & 91.33  & \cellcolor[rgb]{ 1,  .949,  .8}\textbf{90.22 } & 76.67  & 80.11  \\
    Phi3.5-mini-Instruct & 82.50  & 78.44  & 73.00  & 82.89  & 85.56  & 86.67  & 87.78  & 68.44  & 80.11  \\
    Phi3.5-MoE-Instruct & \cellcolor[rgb]{ 1,  .949,  .8}\textbf{90.72 } & \cellcolor[rgb]{ .91,  .898,  .953}\textbf{84.00 } & \cellcolor[rgb]{ 1,  .918,  .898}\textbf{80.78 } & 93.22  & 89.00  & 90.11  & 87.67  & 73.44  & 80.11  \\
    \midrule
    Llama2-7B-Chat & 67.61  & 58.72  & 52.00  & 61.11  & 63.78  & 65.11  & 67.22  & 58.00  & 66.56  \\
    Llama2-13B-Chat & 77.50  & 68.61  & 63.72  & 73.56  & 74.89  & 73.78  & 84.44  & 60.78  & 73.00  \\
    Llama2-70B-Chat & 85.72  & 77.67  & 66.72  & 82.22  & 82.89  & 81.44  & 90.22  & 76.78  & 80.33  \\
    Llama3.1-8B-Instruct & 89.11  & 80.22  & 74.22  & 89.89  & 87.33  & 90.22  & 86.89  & 74.00  & 81.44  \\
    Llama3.1-70B-Instruct & \cellcolor[rgb]{ 1,  .949,  .8}\textbf{93.22 } & \cellcolor[rgb]{ .91,  .898,  .953}\textbf{90.28 } & \cellcolor[rgb]{ 1,  .918,  .898}\textbf{85.89 } & \cellcolor[rgb]{ 1,  .949,  .8}\textbf{92.00 } & \cellcolor[rgb]{ .91,  .898,  .953}\textbf{94.67 } & \cellcolor[rgb]{ 1,  .918,  .898}\textbf{96.00 } & \cellcolor[rgb]{ 1,  .949,  .8}\textbf{91.00 } & \cellcolor[rgb]{ .91,  .898,  .953}\textbf{81.33 } & \cellcolor[rgb]{ 1,  .918,  .898}\textbf{83.33 } \\
    \midrule
    Qwen-7B-Chat & 83.72  & 77.67  & 68.56  & 83.33  & 80.00  & 80.11  & 88.44  & 64.22  & 73.22  \\
    Qwen-14B-Chat & 85.33  & 78.11  & 72.44  & 85.78  & 81.67  & 83.67  & 90.33  & 68.44  & 71.44  \\
    Qwen-72B-Chat & 91.83  & 83.39  & 82.72  & 91.78  & 90.56  & 93.89  & 92.44  & \cellcolor[rgb]{ .91,  .898,  .953}\textbf{82.56 } & 82.78  \\
    Qwen2-7B-Instruct & 83.17  & 74.72  & 72.11  & 84.56  & 82.78  & 81.33  & 81.44  & 71.89  & 78.56  \\
    Qwen2-72B-Instruct & \cellcolor[rgb]{ 1,  .949,  .8}\textbf{95.28 } & \cellcolor[rgb]{ .91,  .898,  .953}\textbf{88.44 } & \cellcolor[rgb]{ 1,  .918,  .898}\textbf{87.06 } & 94.89  & \cellcolor[rgb]{ .91,  .898,  .953}\textbf{93.00 } & \cellcolor[rgb]{ 1,  .918,  .898}\textbf{96.56 } & \cellcolor[rgb]{ 1,  .949,  .8}\textbf{93.00 } & 81.11  & 82.89  \\
    Qwen2.5-7B-Instruct & 89.00  & 85.83  & 80.56  & 90.00  & 90.67  & 89.44  & 83.22  & 76.67  & 82.78  \\
    Qwen2.5-14B-Instruct & 93.56  & 87.11  & 84.28  & 91.78  & 90.33  & 94.11  & 92.00  & 82.33  & \cellcolor[rgb]{ 1,  .918,  .898}\textbf{85.44 } \\
    Qwen2.5-32B-Instruct & 93.39  & 87.94  & 84.56  & \cellcolor[rgb]{ 1,  .949,  .8}\textbf{95.56 } & 92.11  & 95.89  & 92.11  & 80.33  & 81.44  \\
    Qwen2.5-72B-Instruct & 94.67  & 86.06  & 86.50  & 94.78  & 91.67  & 95.22  & 91.33  & 80.67  & 81.44  \\
    \midrule
    GPT-3.5 & 82.83  & 77.11  & 72.89  & 85.00  & 84.78  & 85.89  & 89.11  & 73.78  & 78.78  \\
    GPT-4o mini & 92.44  & 86.83  & 84.89  & 93.33  & 93.33  & 95.11  & 90.44  & 78.44  & 83.89  \\
    GPT-4o & \cellcolor[rgb]{ 1,  .949,  .8}\textbf{93.89 } & \cellcolor[rgb]{ .91,  .898,  .953}\textbf{88.89 } & \cellcolor[rgb]{ 1,  .918,  .898}\textbf{87.67 } & \cellcolor[rgb]{ 1,  .949,  .8}\textbf{94.33 } & \cellcolor[rgb]{ .91,  .898,  .953}\textbf{94.33 } & \cellcolor[rgb]{ 1,  .918,  .898}\textbf{97.67 } & \cellcolor[rgb]{ 1,  .949,  .8}\textbf{93.44 } & \cellcolor[rgb]{ .91,  .898,  .953}\textbf{83.11 } & \cellcolor[rgb]{ 1,  .918,  .898}\textbf{88.89 } \\
    \bottomrule
    \end{tabular}}
    \caption{The experimental results of the CoT prompt-based method on the three types of tasks.}
  \label{cot_results}%
\end{table*}%

\section{Detailed Results of three tasks}
\label{sec:detailed_results}
Tables~\ref{base_results} and \ref{cot_results} present the performance of different models on motivation reasoning, behavior reasoning, and motivation-behavior reasoning tasks. We observe that larger models tend to perform better than smaller models.

\section{Experiment Prompts}
\label{sec:prompts}
Table~\ref{client_prompt} provides a comprehensive overview of the prompts used for model evaluation in M\textsc{otive}B\textsc{ench}, including both the base prompt and the CoT (Chain-of-Thought) prompt. Additionally, Tables~\ref{questioner}, \ref{reviewer}, and \ref{modifier} detail the specific prompts designed for the questioner, reviewer, and modifier roles within our multi-agent question generation and refinement framework.

To ensure high-quality and well-calibrated question modifications, we empirically set the maximum number of modification rounds to 5. The process proceeds iteratively, with the three reviewers assessing different aspects of the generated questions. If no further issues are detected within this threshold, the refinement process is terminated early to maintain efficiency while preserving quality. This structured approach balances thoroughness and computational efficiency, ensuring that the final questions meet our predefined standards.

\begin{table*}[htbp]
  \centering
    \begin{tabular}{|p{\textwidth}|}
    \toprule
    \multicolumn{1}{|c|}{\textbf{Base Prompt for Evaluation}} \\
    \midrule
    The following is a \{Question\_Type\}. You should: \{Type\_Interpretation\}. Carefully read the given question, fully immerse yourself in the role of the character described, and reason based on the information provided. Your answer should rely strictly on the given details.\newline{}Note:\newline{}1. Based on the content of the given question, please infer the most likely answer.\newline{}2. You must select one answer from the given options: "A, B, C, D, E, F" as the most likely choice. Even if the question does not provide sufficient information to determine the correct answer, you should randomly choose one option as your output.\newline{}3. The result can only return **one character without any other explanation**.\newline{}\newline{}Question: \{Question\_Content\}\newline{}\newline{}Options: \{Options\} \\
    \midrule
    \multicolumn{1}{|c|}{\textbf{CoT Prompt for Evaluation}} \\
    \midrule
    The following is a \{Question\_Type\}. You should: \{Type\_Interpretation\}. Carefully read the given question, fully immerse yourself in the role of the character described, and reason based on the information provided. Your answer should rely strictly on the given details.\newline{}Note:\newline{}1. Based on the content of the given question, please think step by step and infer the most likely answer.\newline{}2. You must select one answer from the given options: "A, B, C, D, E, F" as the most likely choice. Even if the question does not provide sufficient information to determine the correct answer, you should randomly choose one option as your output.\newline{}3. Please first think through the question step by step, analyze the reasoning process for the possible answers, and finally output the most likely answer's letter. **The last line of your reply should only contain one character of your final choice.**\newline{}\newline{}Question: \{Question\_Content\}\newline{}\newline{}Options: \{Options\} \\
    \midrule
    \multicolumn{1}{|c|}{\textbf{Illustration}} \\
    \midrule
    \{Question\_Type\}: \{Type\_Interpretation\}\newline{}1. \textbf{Motivational Reasoning Question}: Based on the given scenario and the character's profile, determine the most likely motivation behind the character's behavior.\newline{}2. \textbf{Behavioral Reasoning Question}: Based on the given scenario and the character's profile, determine the most likely behavior the character would take next, given the motivation.\newline{}3. \textbf{Motive\&Behavior Reasoning Question}: Based on the given scenario and the character's profile, determine the most likely motivation the character would develop next and the corresponding behavior would take.
    \newline{}\newline{}
    \{Options\}\\
    To minimize the potential bias caused by the order of options, we will randomize the order of options six times and calculate the average result from these six experiments. Specifically, the order sequences used will be: [1, 2, 3, 4, 5, 6], [6, 5, 4, 3, 2, 1], [3, 1, 6, 5, 4, 2], [2, 3, 5, 6, 1, 4], [5, 4, 1, 2, 6, 3], and [4, 6, 2, 1, 3, 5], ensuring that each option appears in every possible position across the six sequences.\\
    \bottomrule
    \end{tabular}
    \caption{Prompts for evaluation.}
  \label{client_prompt}
\end{table*}
\begin{table*}[htbp]
  \centering
    \begin{tabular}{|p{\textwidth}|}
    \toprule
    \multicolumn{1}{|c|}{\cellcolor[rgb]{ .867,  .922,  .969}\textbf{Prompts of Questioner}} \\
    \midrule
    Consider the four elements of scenario, profile, motivation, and behavior. In a given scenario, a character with a specific profile will perform a certain behavior based on a certain motivation. You are a professional psychologist and sociologist, skilled at creating challenging reasoning questions based on given scenarios to test participants' motivation and behavior reasoning abilities.\newline{}\newline{}**Please create three questions based on the given scenario:**\newline{}\newline{}1. Motivational Reasoning Question: Given a complex scenario, a specific profile, and a given behavior, infer the most likely motivation behind the character's behavior. The question should not contain any direct description related to the predicted motivation.\newline{}2. Behavioral Reasoning Question: Given a complex scenario, a specific profile, and a given motivation, infer the most likely behavior the character will perform based on that motivation. The question should not contain any direct description related to the predicted behavior.\newline{}3. Motive\&Behavior Reasoning Question: This is a more advanced test. The question should only include the complex scenario and the character's profile. Using only the complex scenario and specific profile, infer the most likely motivation the character will have and the corresponding behavior they will perform.\newline{}\newline{}To summarize, all three questions are based on the same story scenario and character profile setup. The motivation reasoning question requires the addition of a behavior in the question stem and asks the participant to infer the motivation for that behavior. The behavior reasoning question requires the addition of a motivation in the question stem and asks the participant to infer the behavior that may result from that motivation. The motivation and behavior reasoning question does not need any additional information and requires the participant to infer both the motivation and behavior of the character based on the given scenario and profile.\newline{}\newline{}**Note:**\newline{}\newline{}1. You will be provided with a simple scenario description. Please rewrite this scenario by correcting any logical inconsistencies, and add relevant details to make the scenario, profile, motivation, and behavior more vivid and complex.\newline{}2. Choose the most appropriate motivation and behavior to create the questions. However, ensure that the motivation and behavior are only related to real human needs, not to any POIs or products in the text.\newline{}3. The three questions are independent of each other and should be answered separately, meaning that each question should only rely on its own stem and not contain any information from the others. Therefore, please ensure that each question has enough rich and complex scenario and profile information to support correct reasoning.\newline{}4. Each question should have only one correct answer, along with five distractors. The distractors must be related to certain parts of the information in the question. Please analyze why each option is correct or incorrect.\newline{}5. The question stem must include irrelevant or redundant information that creates distractions and challenges. This is necessary to ensure each question is challenging. The correct answer must not be explicitly stated in the question. \\
    \bottomrule
    \end{tabular}%
  \caption{Detailed prompts of questioner in the multi-agent framework.}
  \label{questioner}
\end{table*}%
\begin{table*}[htbp]
  \centering
    \begin{tabular}{|p{\textwidth}|}
    \toprule
    \multicolumn{1}{|c|}{\cellcolor[rgb]{ .867,  .922,  .969}\textbf{Prompts of Reviewers}} \\
    \midrule
    You are a strict and discerning psychologist and sociologist, capable of precisely identifying issues in the given behavior and motivation reasoning questions and offering improvement suggestions.\newline{}\newline{}I will provide you with three behavior and motivation reasoning questions. Please evaluate them based on the following aspects:\newline{}\newline{}1. **Reasonableness of the Question Information and Type**: Specifically, all three questions should contain a concrete scenario and character profile. The motivation reasoning question should include additional behavioral information about the character. The behavior reasoning question should include additional motivational information about the character. The motivation and behavior reasoning question should not contain any direct clues about the motivation or behavior.\newline{}2. **Logical Consistency and Reasonableness of the Four-Tuple**: Assess whether, in the given scenario, a character with a specific profile would logically perform the stated behavior based on the provided motivation.\newline{}3. **Sufficiency of Information to Derive the Correct Answer**: Examine whether the information provided in each question is enough to infer the correct answer. If not, suggest modifications to the scenario or character profile to make the information clearer or more comprehensive.\newline{}4. **Challenge and Difficulty of the Question**: Evaluate whether the question presents an appropriate level of difficulty and challenge for the respondent.\newline{}5. **Correct Answer Must Not Be Explicitly Stated**: Ensure that the correct answer does not appear explicitly in the question information and can only be deduced through reasoning steps.\newline{}6. **Clarity and Plausibility of Distractor Options**: Evaluate whether the incorrect options are misleading and whether they correspond to distracting information within the question. If they do not, suggest adding the relevant distracting information or modifying the options.\newline{}7. **Adequate Distractors and Redundant Information**: Ensure that each question includes enough irrelevant or redundant information to make the question challenging, but without disrupting the logic needed to deduce the correct answer.\newline{}8. **Objectivity and Neutrality of the Question**: Ensure that the question is presented in a neutral and objective manner, with no implicit suggestion of the correct answer.\newline{}\newline{}Please provide specific modification suggestions for the question set and give your feedback to the question author in a reasonable tone. Summarize your evaluation into a single paragraph of suggestions.\newline{}\newline{}(All the aspects listed above are of concern, and each reviewer will be asked to focus on different aspects.) \\
    \bottomrule
    \end{tabular}
    \caption{Detailed prompts of reviewers in the multi-agent framework.}
  \label{reviewer}%
\end{table*}%

\begin{table*}[htbp]
  \centering
    \begin{tabular}{|p{\textwidth}|}
    \toprule
    \multicolumn{1}{|c|}{\cellcolor[rgb]{ .867,  .922,  .969}\textbf{Prompts of Modifier}} \\
    \midrule
    Consider the four elements of scenario, character profile, motivation, and behavior. In the given scenario, a character with a specific profile will perform a certain behavior based on a particular motivation. You are a professional psychologist and sociologist, skilled in refining motivation and behavior reasoning test questions and providing relevant suggestions for improvement.\newline{}\newline{}**The specific types of the three questions are as follows:**\newline{}\newline{}1. Motivational Reasoning Question: Based on a complex scenario, a specific character profile, and a given behavior, deduce the most likely motivation behind the character's action. The question should not include any description related to the predicted motivation.\newline{}2. Behavioral Reasoning Question: Based on a complex scenario, a specific character profile, and a given motivation, deduce the most likely behavior the character will perform based on that motivation. The question should not include any description related to the predicted behavior.\newline{}3. Motive\&Behavior Reasoning Question: The question should only include a complex scenario and character profile. This is a more difficult question type, where the respondent must deduce the most likely behavior and corresponding motivation of the character based solely on the scenario and character profile.\newline{}\newline{}In summary, all three questions are based on the same story scenario and character profile settings. For the motivation reasoning question, an additional behavior is given, and the task is to deduce the motivation behind that behavior. In the behavior reasoning question, an additional motivation is given, and the task is to deduce the behavior that would most likely result from that motivation. The motivation and behavior reasoning question, however, does not provide any additional information, requiring the respondent to deduce both the motivation and behavior from the scenario and character profile. It is crucial that the story scenario and character profile in the question are rich enough to support reasoning and lead to the correct answer.\newline{}\newline{}**Specific Requirements:**\newline{}\newline{}1. Carefully consider each suggestion based on the given questions and selectively make reasonable changes to the questions.\newline{}2. Do not delete the distracting information related to the incorrect answers, as this is necessary to ensure the questions remain challenging.\newline{}3. The three questions are independent of each other and are to be answered separately. Respondents should only reason based on the question provided, without seeing any other information. Therefore, ensure that each question has sufficiently rich and complex scenario and character profile information.\newline{}4. After making revisions, analyze each option to determine why it is correct or incorrect. If there are any issues, modify the question again to ensure the uniqueness of the correct answer. \\
    \bottomrule
    \end{tabular}
    \caption{Detailed prompts of modifier in the multi-agent framework.}
  \label{modifier}%
\end{table*}%

\section{Detailed Discussion \& Case Study}
\label{app:cases}
\subsection{General Analysis of GPT-4o’s Errors}
We analyze instances where GPT-4o’s responses deviate from human consensus in motivational and behavioral reasoning tasks. Specifically, for questions where GPT-4o selects incorrect answers, we prompt the model to explain its reasoning behind the chosen option and analyze why the correct answer is appropriate. By systematically comparing incorrect responses with those chosen by human participants, we identify distinct error patterns in GPT-4o’s reasoning. These errors fall into four primary categories:
\begin{itemize}
    \item \textbf{Over-Rationalization, Lacking Emotional Insight.} The model prioritizes logical coherence over emotional or social nuances that influence human decision-making.
    \item \textbf{Weak Logical Precision, Prone to General Assumptions.} The model makes broad generalizations, leading to logical imprecision and a failure to anchor its reasoning in specific contextual evidence.
    \item \textbf{Overly Idealistic, Ignoring Complex Realities.} GPT-4o assumes that idealistic or aspirational actions are more probable, even when contextual evidence suggests a more pragmatic or personally relevant behavior.
    \item \textbf{Lack of Awareness of Behavioral Impact.} The model overlooks the real-world impact of behaviors and motivations, often misjudging the alignment between a character’s intent and plausible actions.
\end{itemize}

\subsection{Detailed Case Study}
\label{app.case}
The Figure~\ref{case_study} provides an example of GPT-4o’s reasoning flaws in a motive and behavior reasoning question. The scenario involves Samantha, a country music artist with deep community ties, who participates in a BBQ collaboration and values authenticity and audience connection. The task asks what behavior she is most likely to exhibit next, considering her motivations.

GPT-4o selects Option D (organizing a community concert to raise funds), justifying this choice by assuming that Samantha’s community ties naturally imply a desire to support local causes through fundraising. However, this reasoning is \textbf{overly idealistic} because there is no contextual evidence suggesting she is motivated by philanthropy. Instead, the scenario emphasizes her \textbf{personal artistic expression and audience engagement}, making composing a song about her BBQ experiences (Option C) the more natural choice.

Furthermore, GPT-4o’s reasoning demonstrates \textbf{weak logical precision} by assuming that community involvement directly translates into organizing fundraising events, rather than considering more direct forms of artistic expression that align with Samantha’s profession. In contrast, human participants recognize that a musician would most naturally process experiences through songwriting rather than logistical event organization.

Beyond this case, our analysis includes multiple other examples illustrating these four error categories. These cases appear in the Tables~\ref{case_1}, \ref{case_2}, \ref{case_3}, \ref{case_4}, \ref{case_5}, where we provide a systematic breakdown of how GPT-4o’s errors manifest across different behavioral reasoning tasks. Refining model training to emphasize emotional intelligence, contextual precision, and realistic behavioral motivations could enhance its alignment with human-like reasoning patterns. 

In Cases 6 and 7, we present examples where GPT-4o arrives at the correct answer through direct answering but produces an incorrect answer when using CoT reasoning. These two cases illustrate two distinct types of errors: (1) the consensus answer is sometimes the one that best fits the context, rather than the one most logically deduced; and (2) the model exhibits a tendency toward overthinking—excessively weighing multiple options, which leads to deviation from the optimal choice.

\begin{table*}[htbp]
  \centering
    \begin{tabular}{|p{\textwidth}|}
    \toprule
    \multicolumn{1}{|c|}{\cellcolor[rgb]{ 1,  .949,  .8}\textbf{Case 1 - Motive\&Behavior Reasoning Question}} \\
    \midrule
    \textbf{Context:} Samantha, a grateful accident survivor, was involved in a severe car crash six months ago. She has since undergone multiple surgeries and intense physical therapy. As a freelance writer before the accident, Samantha now spends much of her free time reading and writing poetry, which she shares occasionally on her personal blog. She has been sharing her recovery journey on social media and feels strongly about using her experience to make a positive impact. Recently, she has gained a lot of traction and connected with many individuals through her posts, deepening her sense of responsibility to those who are still struggling.\newline{}\newline{}\textbf{Question:} What kind of behavior is Samantha most likely to exhibit next, and what is the motivation behind it?\newline{}Options:\newline{}A. Motivation: Cultivating a personal brand; Behavior: Posting artistic photos of her daily life and updates about her writing process.\newline{}B. Motivation: Highlighting the importance of emotional resilience; Behavior: Hosting online webinars focused on mental health strategies.\newline{}C. Motivation: Encouraging community support; Behavior: Organizing small group meetups for accident survivors to share their experiences.\newline{}D. Motivation: Seeking validation from peers; Behavior: Posting emotionally charged poetry on social media to gain likes and shares.\newline{}E. Motivation: Gaining recognition for her journey; Behavior: Collaborating with influencers to promote her story.\newline{}F. Motivation: Finding solace through expression; Behavior: Writing a memoir to reflect on her healing process.\newline{}\newline{}\textbf{Correct Answer: C}\newline{}\textbf{GPT-4o's Answer: B} \\
    \midrule
    \cellcolor[rgb]{ .996,  .957,  .918}\textbf{Analysis:} In this scenario, the appropriate action should be to offer emotional support, not discuss mental health strategies. GPT-4o's choice (hosting a webinar on mental health strategies) focuses on professional methods, which doesn't align with Samantha's current situation. She seeks to inspire others through her personal experiences, not teach strategies. GPT-4o's reasoning is too theoretical, lacking the empathy and life experience humans use to understand motivations. Humans, considering Samantha’s struggles, would focus on actions that resonate with her personal healing process, such as sharing her story to help others. \\
    \bottomrule
    \end{tabular}
    \caption{Case 1 on a Motive\&Behavior Reasoning Question.}
  \label{case_1}%
\end{table*}%

\begin{table*}[htbp]
  \centering
    \begin{tabular}{|p{\textwidth}|}
    \toprule
    \multicolumn{1}{|c|}{\cellcolor[rgb]{ 1,  .949,  .8}\textbf{Case 2 - Motive\&Behavior Reasoning Question}} \\
    \midrule
    \textbf{Context:} James, a seasoned stockbroker specializing in tech and software stocks, has recently noticed a growing interest among younger investors in sustainable and socially responsible investments. Despite his initial skepticism, he recognizes the potential effects of this trend on his career. Additionally, he faces increasing competition from newer, digitally-savvy brokers capitalizing on this shift. Furthermore, he has come across various articles detailing the increasing demand for sustainable investments.\newline{}\newline{}\textbf{Question:} Given these circumstances, what kind of behavior is James most likely to exhibit next, and what could be the motivation behind it?\newline{}\newline{}\textbf{Options:}\newline{}A. He will develop a marketing strategy aimed at promoting sustainable tech stocks, driven by the desire to connect with a younger audience interested in socially responsible investments.\newline{}B. He will partner with a fintech firm specializing in sustainable investments, motivated by the necessity to broaden his service offerings and enhance client retention.\newline{}C. He will begin writing articles for finance magazines, driven by the ambition to share his insights on the importance of sustainable investing among his peers.\newline{}D. He will initiate a webinar series focusing on sustainable investment trends, motivated by the goal of showcasing his expertise and engaging with potential clients.\newline{}E. He will host social events for potential investors, driven by the intention to foster relationships and promote discussions around sustainable investing.\newline{}F. He will create a newsletter highlighting sustainable investment options, motivated by the aim of educating clients about emerging trends in the market.\newline{}\newline{}\textbf{Correct Answer: A}\newline{}\textbf{GPT-4o's Answer: D} \\
    \midrule
    \cellcolor[rgb]{ .996,  .957,  .918}\textbf{Analysis:} When faced with new market trends, humans typically prioritize directly addressing market demands and customer interests. For example, the growing interest of young investors in sustainable investments leads to a marketing strategy tailored to this group, which aligns with real market needs. GPT-4o tends to suggest that James might showcase his expertise through a webinar, but this motivation focuses more on "self-promotion" rather than directly responding to market demands or attracting a specific group, failing to address the competitive pressures and market changes. \\
    \bottomrule
    \end{tabular}
    \caption{Case 2 on a Motive\&Behavior Reasoning Question.}
  \label{case_2}%
\end{table*}%

\begin{table*}[htbp]
  \centering
    \begin{tabular}{|p{\textwidth}|}
    \toprule
    \multicolumn{1}{|c|}{\cellcolor[rgb]{ 1,  .949,  .8}\textbf{Case 3 - Motive\&Behavior Reasoning Question}} \\
    \midrule
    \textbf{Context:} Yara, a new mother with a newborn baby girl who has a history of allergies, recently dined at a café that provided detailed ingredient lists and used allergen-safe cooking methods. She was satisfied with the café's attention to allergen management and its ability to cater to her needs. Yara is also known to actively participate in community groups focused on managing allergies in children.\newline{}\newline{}\textbf{Question:} In this scenario, what is Yara most likely to do next, and what is her primary motivation?\newline{}\newline{}\textbf{Options:}\newline{}A. Sharing her experience with others, motivated by her commitment to helping the allergy community.\newline{}B. Thanking the café staff, motivated by appreciation for their allergen-safe practices.\newline{}C. Researching other restaurants, motivated by a desire for variety in dining options.\newline{}D. Leaving a negative review elsewhere, motivated by frustration over previous dining challenges.\newline{}E. Avoiding dining out altogether, motivated by concerns about public allergens.\newline{}F. Offering advice to another parent in the café, motivated by her interest in parenting discussions.\newline{}\newline{}\textbf{Correct Answer: B}\newline{}\textbf{GPT-4o's Answer: A} \\
    \midrule
    \cellcolor[rgb]{ .996,  .957,  .918}\textbf{Analysis:} GPT-4o's analysis overlooks emotion-driven behavior by focusing on Yara's rational and altruistic motives, assuming she would share her experience to help others. This perspective ignores the possibility of a direct emotional response, such as expressing gratitude for the cafe service. Furthermore, GPT-4o overinterprets Yara's background in community management, predicting that her actions would be more focused on helping others or sharing experiences, rather than simply thanking the staff. In contrast, humans are more likely to recognize that, despite Yara’s involvement in the community, her immediate interaction with the cafe staff would be influenced by her emotional response, such as gratitude, fitting the context of the situation. \\
    \bottomrule
    \end{tabular}
    \caption{Case 3 on a Motive\&Behavior Reasoning Question.}
  \label{case_3}%
\end{table*}%

\begin{table*}[htbp]
  \centering
    \begin{tabular}{|p{\textwidth}|}
    \toprule
    \multicolumn{1}{|c|}{\cellcolor[rgb]{ 1,  .949,  .8}\textbf{Case 4 - Motive\&Behavior Reasoning Question}} \\
    \midrule
    \textbf{Context:} Samantha owns a travel agency that specializes in personalized service and unique travel experiences. Recently, she’s been thinking about ways to make her business more environmentally friendly and believes that adopting sustainable practices could also boost her agency's reputation. During a meeting with her accountant, Mark, they reviewed various financial strategies to implement her ideas. Although the agency already has basic recycling and uses digital communication to reduce waste, Samantha is determined to make a bigger impact in the competitive travel market.\newline{}\newline{}\textbf{Question}: What kind of behavior is Samantha most likely to exhibit next, and what is the motivation behind it?\newline{}\newline{}\textbf{Options:}\newline{}A. Announcing a new program to contribute most profits to local environmental projects, motivated by a desire to build the agency’s reputation for community involvement.\newline{}B. Rushing to install solar panels on all properties without detailed cost planning, motivated by an urgent need to show visible commitment to sustainability.\newline{}C. Delaying new projects until further discussions with stakeholders, motivated by caution about potential financial risks.\newline{}D. Expanding the recycling program to engage customers in eco-friendly actions, motivated by a focus on community-based solutions.\newline{}E. Launching a promotional campaign about the agency’s past sustainable practices, motivated by the desire to draw media attention.\newline{}F. Organizing workshops for employees on sustainable practices, motivated by a goal to enhance internal awareness.\newline{}\newline{}\textbf{Correct Answer: D}\newline{}\textbf{GPT-4o's Answer: F} \\
    \midrule
    \cellcolor[rgb]{ .996,  .957,  .918}\textbf{Analysis:} GPT-4o overlooked the emphasis on "enhancing market competitiveness" in the question and focused excessively on the superficial logic of "sustainability." However, the purpose of sustainability is to enhance market competitiveness, and merely raising internal employees' awareness does not contribute to improving market competitiveness. \\
    \bottomrule
    \end{tabular}%
    \caption{Case 4 on a Motive\&Behavior Reasoning Question.}
  \label{case_4}%
\end{table*}%

\begin{table*}[htbp]
  \centering
    \begin{tabular}{|p{\textwidth}|}
    \toprule
    \multicolumn{1}{|c|}{\cellcolor[rgb]{ 1,  .949,  .8}\textbf{Case 5 - Motive\&Behavior Reasoning Question}} \\
    \midrule
    \textbf{Context:} Fiona, a young woman working as an editor for a prestigious publishing house, lives alone in a vibrant urban neighborhood known for its diverse cultures. One afternoon, after a challenging week at her job, she decides to visit Bella Vita, a charming pizzeria in a more upscale area. Bella Vita is famous for its delicious pizzas and warm Italian atmosphere, complete with nostalgic music and friendly staff. As she sits by the window, enjoying the sunlight, she finds herself laughing softly, and her exhaustion starts to fade. The cozy ambiance surrounds her, bringing her feelings of comfort and joy. Fiona highly values her personal time, often enjoying these quiet moments for reflection and renewal, while also cherishing fond memories of family gatherings at similar Italian restaurants.\newline{}\newline{}\textbf{Question:} As she listens to the familiar tunes and observes families enjoying meals together, based on what motivation is she most likely to exhibit what behavior next?\newline{}\newline{}\textbf{Options:}\newline{}A. Reach out to a friend to share her experience, motivated by her desire for emotional connection.\newline{}B. Jot down her thoughts about the atmosphere, driven by her need for self-expression.\newline{}C. Plan to revisit the restaurant with her family, inspired by her longing for shared memories.\newline{}D. Explore other nearby restaurants, motivated by her curiosity about the local dining scene.\newline{}E. Compliment the staff for their service, reflecting her appreciation for kindness and hospitality.\newline{}F. Take a photograph to post online, motivated by her interest in sharing aesthetic moments with others.\newline{}\newline{}\textbf{Correct Answer: C}\newline{}\textbf{GPT-4o's Answer: B} \\
    \midrule
    \cellcolor[rgb]{ .996,  .957,  .918}\textbf{Analysis:} GPT-4o tends to over-rely on explicit textual details while overlooking implicit behavioral tendencies and deeper emotional motivations. For instance, it often focuses on directly stated traits in the prompt (e.g., “She is an editor, so she may prefer writing”) and limits its reasoning to surface-level information, ignoring how emotions like nostalgia might influence behavior. In contrast, humans naturally consider the emotional undertones within a situation, such as how a familial atmosphere may evoke empathy and drive planning. Additionally, GPT-4o primarily relies on explicit contextual details to infer motivations, whereas humans are more sensitive to subtle emotional cues embedded in the broader scenario, allowing for a more nuanced understanding of behavior. \\
    \bottomrule
    \end{tabular}
    \caption{Case 5 on a Motive\&Behavior Reasoning Question.}
  \label{case_5}%
\end{table*}%

\input{cot_cases}

\end{document}

%% file: cot_cases.tex
\definecolor{correct_case_color}{RGB}{200, 235, 190} 
\definecolor{wrong_case_color}{RGB}{255, 225, 180} 

\newpage
\onecolumn
\begin{tcolorbox}[
  title= Case 6 - A wrong CoT response from  Motive\&Behavior Reasoning Question,
  breakable,   
  fonttitle=\bfseries,
  enhanced,                        
  colback=wrong_case_color!5,
  colbacktitle=wrong_case_color,
  coltitle=black,
  colframe=wrong_case_color!80!black,
  coltext=black,
  boxrule=0.6pt,
  arc=2mm
]
\textbf{Context:} Alex, a renowned nature photographer known for blending stunning natural landscapes with subtle biblical imagery, recently had a successful gallery opening where he explained the symbolic meanings behind his photographs to a diverse group of attendees. The gallery featured a mix of serene landscapes and powerful images that highlighted environmental degradation. Public debates about the role of art in environmental activism have gained momentum, and several environmental groups have reached out to him for collaborations.\newline{}\newline{}\textbf{Question:} Given his background and the current interest in his work, what kind of behavior is this character most likely to exhibit next, and what is the motivation behind it?\newline{}\newline{}\textbf{Options:}\newline{}A. Motivation: To educate others about environmental issues and inspire action; Behavior: Create a documentary series showcasing the impact of human activity on natural habitats.\newline{}B. Motivation: To explore new artistic techniques; Behavior: Experiment with digital art forms and virtual reality installations.\newline{}C. Motivation: To gain more followers; Behavior: Increase his social media presence by posting daily updates and behind-the-scenes content.\newline{}D. Motivation: To document historical sites; Behavior: Photograph ancient ruins and historical landmarks around the world.\newline{}E. Motivation: To support ongoing environmental initiatives; Behavior: Apply for grants and sponsorships to fund environmental organizations.\newline{}F. Motivation: To achieve academic recognition; Behavior: Publish a research paper on the psychological effects of nature photography.\newline{}

\textbf{GPT-4o's CoT Response:} Let’s think step by step. Alex is a renowned nature photographer whose work blends natural landscapes with subtle biblical imagery. At a recent successful gallery opening, he engaged with a diverse audience, explaining the symbolic layers of his art. The gallery included both serene scenes and stark depictions of environmental degradation, revealing a dual interest in beauty and social commentary. Following this, environmental groups reached out to him for collaboration, and public conversations around art’s role in environmental activism have intensified.

Option A suggests that Alex would be motivated to educate and inspire through a documentary series on human impact on nature. This direction is clearly consistent with his existing themes and his engagement with the public at the gallery. A documentary series would allow him to reach a broad audience and continue the conversation beyond the gallery space. However, this step involves a significant shift in medium and production scale, and there's no direct indication in the prompt that Alex is planning to enter the world of film.

Option B presents an artistic motive: the desire to explore new techniques, leading to experimentation with digital art or VR. While creative growth is always plausible for an artist, the narrative centers more on his message-driven work and recent interaction with social causes than on stylistic exploration.

Option C proposes that Alex would focus on increasing his social media activity. While this could help amplify his message, there's no strong emphasis in the prompt on building a following or digital branding. His engagement seems more oriented toward in-person discussion and deeper, symbolic expression than toward constant online sharing.

Option D imagines him turning to photograph historical landmarks. This deviates from his established pattern of combining nature and symbolic commentary. The shift toward historical documentation would represent a departure from the environmentally focused narrative presented in the gallery.

Option E centers on Alex supporting environmental initiatives by seeking funding for organizations. Given that several environmental groups have already approached him, this step fits naturally. It doesn’t require him to change artistic direction but builds on current momentum, translating his influence into tangible support for environmental causes. It reflects a practical, direct action that matches both his thematic work and recent developments—public interest, collaboration opportunities, and alignment with broader environmental movements.

Option F introduces an academic goal: publishing a research paper on the psychological effects of nature photography. Although interesting, this behavior is more typical of a scholar than a photographer actively engaging in public dialogue and visual storytelling.

While Option A maintains strong alignment with Alex’s messaging and public engagement, Option E presents a next step that grows organically out of his recent interactions and the context around his work. The outreach from environmental organizations provides an immediate channel through which Alex can have real-world impact. Applying for grants and sponsorships to support these causes allows him to remain grounded in his artistic identity while leveraging his success to assist larger efforts. For that reason, the most likely behavior is found in Option E: motivated by a desire to support ongoing environmental initiatives, Alex would take action to fund environmental organizations.

The answer is \textbf{E}.\\

\textbf{GPT-4o's Direct Response: A} \\

\textbf{Correct Answer: A}
\end{tcolorbox}

\begin{tcolorbox}[
  title= Case 7 - A wrong CoT response from  Motivation Reasoning Question,
  breakable,   
  fonttitle=\bfseries,
  enhanced,                        
  colback=wrong_case_color!5,
  colbacktitle=wrong_case_color,
  coltitle=black,
  colframe=wrong_case_color!80!black,
  coltext=black,
  boxrule=0.6pt,
  arc=2mm
]
\textbf{Context:} John, a well-known music producer, has built a career on discovering and mentoring new talent. Recently, he has taken an interest in Alex, a young musician who shares his original songs online. Although Alex is relatively unknown, John has offered to produce his first album and introduce him to key industry figures. John spends significant time helping Alex, even adjusting his own schedule to accommodate Alex’s recording sessions.\newline{}\newline{}\textbf{Question:} What is the most likely motivation behind John's actions?\newline{}\newline{}\textbf{Options:}\newline{}A. John believes Alex has the potential to become a successful artist and wants to support his growth.\newline{}B. John is looking for a new business opportunity to expand his production company.\newline{}C. John finds mentoring young talent rewarding and enjoys helping them develop their careers.\newline{}D. John wants to boost his own reputation by working with emerging talent.\newline{}E. John sees Alex as someone who can take on a leadership role in his company in the future.\newline{}F. John is interested in using Alex’s style to try out new music production techniques.\newline{}

\textbf{GPT-4o's CoT Response:} Let’s think step by step. John is a well-known music producer with a track record of discovering and mentoring new talent. His recent interest in Alex, a relatively unknown musician sharing original songs online, seems to follow that same pattern. What stands out is how involved John becomes—offering to produce Alex’s first album, facilitating industry connections, and even shifting his own schedule to prioritize Alex’s recording sessions. This level of investment suggests that John’s motivation goes beyond mere business or experimentation.

Option A posits that John believes in Alex’s potential and wants to support his growth. This interpretation is supported by the facts: John identifies promising talent, and he’s clearly devoting time and resources to help Alex succeed. This makes sense considering his past behavior in discovering and nurturing artists. There’s a personal belief in Alex’s capabilities implied in the way John alters his own routine to assist him. However, this motivation is focused on belief in Alex specifically—it centers on John’s perception of Alex’s individual promise.

Option B suggests a more commercial motive—looking for a new business opportunity. While that could be a side benefit, the narrative emphasizes John’s personal involvement and adjustment of his schedule, which goes beyond what one might expect if the motivation were purely business. There’s no mention of contracts, profit margins, or expansion strategies in the scenario.

Option C presents the idea that John finds mentoring itself fulfilling. Given his history of working with new talent, and the fact that he is willingly investing time and energy into helping Alex—without any immediate mention of financial return—this motivation seems to resonate. John seems to derive intrinsic value from seeing young musicians grow under his guidance. The fact that he helps Alex before Alex has any fame or proven market success suggests that his actions are driven by personal satisfaction in mentoring rather than strategic gain.

Option D implies a self-serving goal: boosting his own reputation. While working with fresh talent can certainly improve a producer’s image, the prompt doesn’t mention any public promotions or efforts by John to capitalize on this association. The focus is on what John is doing for Alex, not what he’s doing to elevate himself.

Option E introduces a more organizational, long-term vision—seeing Alex as a potential leader within John’s company. That’s a plausible corporate motivation, but there’s no support for it in the text. There’s no mention of company roles, structure, or internal planning.

Option F claims John is interested in Alex’s style for experimenting with new techniques. While this could be true in part, again, the depth of John’s commitment—producing an entire album and making industry introductions—suggests something more than just a testing ground for production methods.

At this point, both A and C seem to fit well with what we know. It’s entirely plausible that John sees something special in Alex and wants to help him grow—that would explain why he’s gone out of his way, adjusted his schedule, and taken such an active role. It’s the kind of thing someone does when they genuinely believe in a person’s potential. That belief could easily be the primary force behind his actions.

But then again, there’s something about the consistency of John’s behavior that keeps pointing back to the idea of mentorship itself being the real motivator. This isn’t a one-time gesture—John has built his career around nurturing talent. Maybe it’s not just about Alex, but about what Alex represents: another opportunity for John to do what he genuinely enjoys. The time, the effort, the willingness to step in so early—it all feels like the kind of commitment someone makes when they’re driven by something deeper than belief in one person. Maybe it started with recognizing potential, but it continues because the process is meaningful in itself.

After going back and forth, it becomes harder to see the belief in Alex alone as the full explanation. That belief may have been the spark, but it’s the enjoyment of mentorship—the reward John gets from helping someone grow—that seems to sustain his involvement. In the end, Option C feels like the clearest answer: John finds mentoring young talent rewarding and enjoys helping them develop their careers.

The answer is \textbf{C}.\\

\textbf{GPT-4o's Direct Response: A} \\

\textbf{Correct Answer: A}
\end{tcolorbox}